%% file: main.tex
\documentclass[10pt,twocolumn,letterpaper]{article}

\usepackage{cvpr}              %

\input{utils/preamble}

\definecolor{cvprblue}{rgb}{0.21,0.49,0.74}
\usepackage[pagebackref,breaklinks,colorlinks,citecolor=cvprblue]{hyperref}

\def\paperID{5075} %
\def\confName{CVPR}
\def\confYear{2024}

\title{\nameMethod: A Modern PBR Materials Dataset}

\author{Giuseppe Vecchio\\
University of Catania\\
{\tt\small giuseppe.vecchio@phd.unict.com}
\and
Valentin Deschaintre\\
Adobe Research\\
{\tt\small deschain@adobe.com}
}

\let\oldtwocolumn\twocolumn
\renewcommand\twocolumn[1][]{%
    \oldtwocolumn[{#1}{
        \begin{center}
            \captionsetup{type=figure}
            \includegraphics[width=0.95\textwidth]{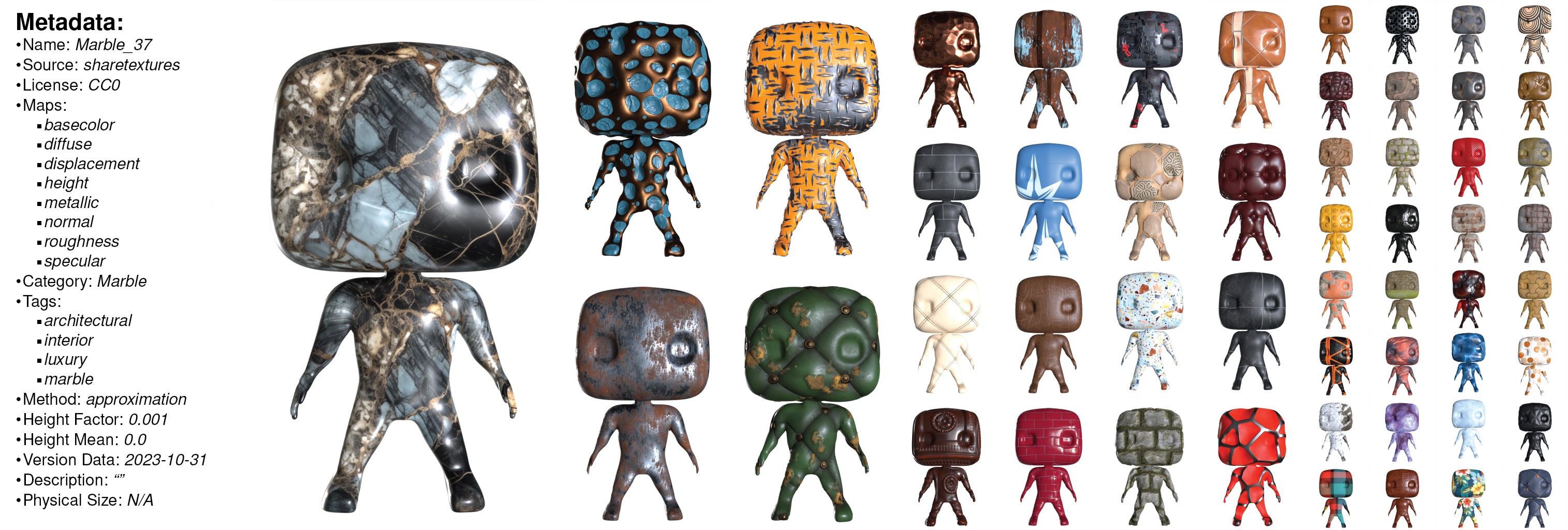}
            \captionof{figure}{\textbf{Samples of materials from the dataset.} Our dataset contains $4,069$ high-quality, 4K, tileable materials with permissive licences. Each material is augmented, rendered and supplemented by metadata containing its origin, tags, categories, method of creation, and more.}
            \label{fig:thumbnail}
        \end{center}
    }]
}

\begin{document}
\maketitle
\input{sec/0_abstract}    
\input{sec/1_intro}
\input{sec/2_related}
\input{sec/3_dataset}

\input{sec/5_results}
\input{sec/6_discussion}
\input{sec/7_ack}

{
    \small
    \bibliographystyle{ieeenat_fullname}
    \bibliography{main}
}

\end{document}

%% file: utils/preamble.tex
\usepackage[dvipsnames]{xcolor}
\usepackage{pgfplots}
\usepackage{pgf-pie}
\pgfplotsset{compat=1.18}
\usepackage{gensymb}
\usepackage{multirow}
\usepackage{rotating}

\newcommand{\suppmat}[1]{\color{black}#1}

\newcommand{\nameMethod}{{\mbox{MatSynth}}\xspace}

%% file: sec/0_abstract.tex
\begin{abstract}
We introduce \nameMethod, a dataset of $4,000+$ CC0 ultra-high resolution PBR materials. Materials are crucial components of virtual relightable assets, defining the interaction of light at the surface of geometries. Given their importance, significant research effort was dedicated to their representation, creation, and acquisition. However, in the past 6 years, most research in material acquisition or generation relied either on the same unique dataset, or on company-owned huge library of procedural materials. With this dataset, we propose a significantly larger, more diverse, and higher resolution set of materials than previously publicly available. We carefully discuss the data collection process and demonstrate the benefits of this dataset for material acquisition and generation applications. The complete data further contains metadata with each material's origin, license, category, tags, creation method, and, when available, descriptions and physical size, as well as 3M+ renderings of the augmented materials, in 1K, under various environment lightings. The MatSynth dataset is released through the project page at: \url{https://www.gvecchio.com/matsynth}. 
\end{abstract}

%% file: sec/1_intro.tex
\section{Introduction}
\label{sec:intro}
Materials are a key component of the rendering pipeline, describing the appearance of objects, for example in terms of colors, shininess or metallicity. The design of high-quality virtual materials is a complex endeavor, usually requiring significant artistic expertise~\cite{pbr_book}. To reproduce existing objects and surfaces, material acquisition and generation have been active research fields for decades~\cite{Guarnera16}. More recently, the field turned to Machine Learning to tackle the ill-posedness of few images (lightweight) material acquisition~\cite{Li17, Deschaintre18}, and to explore new opportunities in material generation\cite{Guo20, Zhou22}. Since 2018, most of the field has relied on a single dataset~\cite{Deschaintre18} or, when possible, on company-owned large material libraries~\cite{Source:2023, Martin22, Zhou22, vecchio2023controlmat}. As company owned libraries are typically orders of magnitude bigger than what is publicly available, they enable very high quality results and experiments with higher resolutions. These materials are, however, not available to the community, underlining the need for larger, more diverse, public datasets to enable further research. Such material data are also particularly useful for synthetic data generation, for example, for object acquisition~\cite{li2018learning, Deschaintre21}, material-based selection~\cite{sharma23}, and numerous applications further from material understanding and authoring~\cite{song2023synthetic, infinigen2023infinite, vecchio_thesis}.

To bridge the gap between private and public datasets, we present \nameMethod, a dataset of $4,069$ tileable materials, available in 4K. Alongside the material properties themselves, we render $3,417,960$ renderings with various scales cropping, rotation, and environment illuminations to facilitate future learning. The materials all come with permissive licences and were sourced from various libraries~\cite{PolyHeaven, ShareTextures, AmbientCG, CGBookCase, TextureCan} and checked for duplication among newly gathered data and against previous datasets~\cite{Deschaintre18}. We further ensure that all materials are represented by the same set of parameters and that they are tileable. In particular, each material is represented by 7 textures/maps: Base color, Diffuse, Normal, Height, Roughness, Metallic and Specular, ensuring compatibility with both Diffuse/Specular (e.g.~\cite{Deschaintre18}) and Base Color/Metallic (e.g.~\cite{Martin22}) workflows. Some materials in our dataset also have an opacity map to handle, for example, surfaces with holes. Finally, we process the previously available dataset to enrich it with metallic, base color, and height properties, facilitating their combinations for a total of $\sim6,000$ materials available for research, with clear licenses attached. 

To validate the benefit of these additional data, we train recent methods with publicly available implementation and evaluate both qualitatively and quantitatively the results quality for different single image material capture methods~\cite{Deschaintre18, vecchio2021surfacenet} and material generation~\cite{vecchio2023matfuse}, showing that the additional data enables higher quality acquisition and improved diversity in the generation results.

In summary, we gather a new material dataset with the following properties and evaluate its impact on material related tasks:
\begin{itemize}
    \item $4,069$ unique 4K PBR materials with permissive licences.
    \item $683,592$ augmented materials with rotations and crops
    \item $3,417,960$ renderings of the augmented materials under various illumination.
    \item Associated metadata.
\end{itemize}

%% file: sec/2_related.tex
\section{Related Work}
\label{sec:related}
We describe the main existing material dataset and related work leveraging it. We also discuss material acquisition/generation approaches using company-owned data or unsupervised learning.

\subsection{Existing dataset}

The only large-scale, publicly available, spatially-varying material dataset contains 1628 distinct, pre-augmentation materials~\cite{Deschaintre18} with a research-only license. 512x512~\cite{Deschaintre19} and 2K~\cite{Deschaintre20} versions of the materials were later published. It is based on random variations from a set of 155 procedural materials gathered from Allegorithmic Substance Share in 2017, distributed in 9 categories: paint(6), plastic(5), leather(13), metal(35), wood(23), fabric(6), stone(25), ceramic tiles(29), ground(13). 
Other datasets used in the field contain 100 and 62 measured BRDFs~\cite{MERL, Dupuy2018Adaptive}, 170 SVBRDFs licensed from VRay\cite{Li17, VRay:2023} or more recently 211 materials from online libraries~\cite{Guo23}. Adobe 3D Assets~\cite{Source:2023}'s $11,000+$ procedural assets have been used to unlock significant quality improvement in the methods that could benefit from it for acquisition, generation, retrieval and captioning~\cite{Zhou22, Martin22,vecchio2023controlmat, deschaintre23_visual_fabric}. This data is however difficult to license for Machine Learning, limiting its use to company internal users.
In contrast, we propose a systematic gathering of non-duplicate, tileable, publicly available materials with permissive licenses from various online libraries, assembling $4,000$+ materials and their metadata, significantly increasing the variety and amount of ressources available for material-related research.

\subsection{Material acquisition}

A typical use of synthetic material data is acquisition method training. Given one or a few images, these methods aim at recovering material properties. As the complete acquisition of a material behaviour typically requires hundreds to thousands of measurements~\cite{Guarnera16}, few images acquisition is a particularly ill-posed problem, which can strongly benefits from machine learning. In the past few years many methods were proposed to recover materials from one~\cite{Li17, Deschaintre18, HLaware, vecchio2021surfacenet, Martin22, Guo23, vecchio2023controlmat} or a few images~\cite{Deschaintre19, Gao19,Deschaintre20, Guo20, fischer2022metappearance}, entirely relying on synthetic data for training. While most of the field relied on synthetic data, Zhou et al.~\cite{xilong2021ASSE} proposed to use a mix of synthetic and real data, and \citet{henzler2021neuralmaterial} leverage a weakly supervised training combined with test-time fine tuning for stationary material acquisition. Some of these work use the large scale Adobe 3D Assets library~\cite{Source:2023, Martin22, vecchio2023controlmat} showing high quality results, suggesting that access to larger public materials datasets will be critical for further progress in the material acquisition field, for both higher quality, higher diversity and higher resolution results.

\subsection{Material generation}
Recent years saw fast quality improvement in generative tasks~\cite{stylegan2}, inspiring research around material generation. In generative workflow in particular, data has been a key enabler for higher quality and diversity~\cite{ramesh2022hierarchical, brown2020language}. In the context of materials, this lead generative research to rely on private large material libraries~\cite{Source:2023}, making contributions to the field challenging for academics or companies which do not own such data. Generative models for materials leveraging supervised GANs were first proposed~\cite{Guo20, zhou2022tilegen}. More recently, a few approaches tackled the problem using text and image conditional diffusion models~\cite{vecchio2023controlmat, vecchio2023matfuse} or tried to simplify the required data by relying solely on weakly supervised flash images during the training~\cite{Zhou23b}. 
Aside from two~\cite{vecchio2023matfuse, Zhou23b}, all related works discussed in this paragraph were trained on private data~\cite{Source:2023}. A concurrent work~\cite{text2mat_pg} follows a similar data gathering approach as ours, but do not make their dataset available to this day, with no plans for it announced. We believe that our dataset is a siginificant step towards reducing the gap between publicly and privately available material datasets, facilitating future research.

An orthogonal research direction, focused on procedural material generation~\cite{guerrero2022matformer, hu2023gen} requires access to complete procedural materials, which is out of our dataset's scope.

\subsection{Data generation}
Aside from the flat material domain, different synthetic datasets, mostly of interior scenes~\cite{roberts:2021, zhu2022learning, sharma23, li2021openrooms}, have been proposed to enable further research for tasks such as inverse rendering and intrinsic image decomposition ~\cite{li2020inverse,zhu2022irisformer}, relighting~\cite{griffiths2022outcast}, guided assets retrieval~\cite{Yan:2023:PSDR-Room} or material similarity and selection~\cite{Perroni-Scharf_2022_CVPR, sharma23}. More generally, synthetic data has been an important resources for various world generation approaches~\cite{song2023synthetic, infinigen2023infinite, vecchio_thesis}, providing ground truth properties to bootstrap trainings, generalising to real images. Larger materials datasets availability will significantly simplify the creation of variations of existing scene datasets, increasing the diversity of appearances, for example through similar material retrieval or random assignment, as done recently with private material libraries~\cite{Perroni-Scharf_2022_CVPR, sharma23}.

%% file: sec/3_dataset.tex
\section{The Dataset}
\label{sec:dataset}
The MatSynth dataset is designed to support modern learning-based techniques for a variety of material-related tasks including, but not limited to, material acquisition, material generation, and synthetic data generation, e.g., for retrieval or segmentation. It improves on existing available datasets by providing a large collection ($\sim2.5$ times larger than previously available datasets) of non-duplicate, high-quality, high-resolution realistic materials alongside useful metadata. 
Each material in the dataset is represented by a set of reflectance maps (basecolor, diffuse, normal, height, roughness, metallic, and specular), complemented by fine-grained annotations including the source and link to the asset, its license, and original author (when available), tags, the creation method, whether is stationary or not, a timestamp for versioning, a description (when available), and the material physical size (when available).

In this section, we describe the \nameMethod dataset, its gathering and analyse its content.

\subsection{Materials Collection}
\label{sec:data_collection}
To gather the \nameMethod dataset we extensively collect data from multiple online sources operating under the CC0 and CC-BY licensing framework. This collection strategy allows to capture a broad spectrum of materials, from commonly used ones to more niche or specialized variants while guaranteeing that the data can be used for a variety of usecases. 
Materials under CC0 license were collected from AmbientCG~\cite{AmbientCG}, CGBookCase~\cite{CGBookCase}, PolyHeaven~\cite{PolyHeaven}, ShateTexture~\cite{ShareTextures}, and TextureCan~\cite{TextureCan}. The dataset also includes limited set of materials from the artist Julio Sillet, distributed under CC-BY license.
We collected over 6000 materials which we meticulously filter to keep only tileable, 4K materials. This high resolution allows us to extract many different crops from each sample at different scale for augmentation. Additionally, we discard blurry or low-quality materials (by visual inspection). The resulting dataset consists of 3736 unique materials which we augment by blending semantically compatible materials (e.g.: snow over ground). In total, our dataset contains 4069 unique 4K materials. We ensure that these new materials are not duplicates of those previously available in the existing material dataset~\cite{Deschaintre18}.

\subsection{Data Annotations}
The dataset is composed of material maps (Basecolor, Diffuse, Normal, Height, Roughness, Metallic, Specular and, when useful, opacity) and associated renderings under varying environmental illuminations, and multi-scale crops.
We adopt the OpenGL standard for the Normal map (Y-axis pointing upward). The Height map is given in a 16-bit single channel format for higher precision.

In addition to these maps, the dataset includes other annotations providing context to each material: the capture method--whether it is photogrammetry, procedural generation, or approximation--; list of descriptive tags; source name (website); source link ; licensing and a timestamps for eventual future versioning. For a subset of materials, when the information is available, we also provide the author name (387), text description (572) and a physical size, presented as the length of the edge in centimeters (358).

\subsection{Data Processing}
We carefully evaluate and select the collected data. We first visually inspect each material, looking for evident defects in the maps (e.g.: blurred maps, baked in shadows or highlights, unrealistic reflectance properties). The remaining materials have been automatically checked using CLIP~\cite{clip} embeddings and contrastive prompts as described in~\cite{wang2023exploring}. In particular, we evaluated each rendered material for quality with the following contrastive pairs: quality (high-quality/low-quality), sharpness (sharp/blurry), noisiness (clean/noisy), and realism (natural/synthetic).

We check for materials duplication by cross comparing the CLIP~\cite{clip} embeddings of all materials in the dataset, flagging values close to $1$ as possible duplicate. Each duplicate candidate was then manually reviewed for false positives before being removed. 

We ensure that all materials respect the assumed y-axis pointing upward by differentiating the height map into a normal map and counting the number vectors with a discordant y component. For materials with a discordant count above 30\%, we manually check the normal map and invert the y component when necessary. We perform a final visual inspection of the renders to ensure a correct normal orientation for all materials in the dataset.

Finally, we automatically compute an estimate of the correct displacement factor for each material, by optimizing a displacement factor parameter to match the normal vectors derived from the scaled height map to the ground truth normal map. This ensures a good match between the normal and the displacement scale when rendering. We provide the correctly scaled height map, as well as the original and the scaling parameters (scale factor and mid-value compatible with the Blender ``Displacement'' node).

\subsection{Data Augmentation}
\input{figures/blend_samples/blend_samples}
\input{figures/render_samples/render_samples}

\noindent\textbf{Material blending.}
We further enrich the dataset by blending semantically compatible materials. First, we define a set of compatible classes in terms of base layer (the base of the material) and top layer (the material to be layered on top of it). Second, we compute a blending mask based on the combination of the two height maps. To do so, we randomly sample a threshold between the minimum and maximum values of the base layer material, we shift the top material height mean to the sampled threshold, and intersect the two height maps taking the material with the max height value for each pixel. We finally blur the blending mask with a 9 pixels Gaussian kernel to smooth the transition between the two materials. Each map of the blended material is computed as $M = M_{base} \times mask + M_{top} \times (1 - mask)$. We include the blending mask for each blended material in the dataset.
Fig~\ref{fig:blend_samples} shows examples of blended materials, the original materials and the blending masks.
We generate a total of 332 blends. \suppmat{The full list of class pairs used for blending, and the number of blends, is provided in the supplemental material.}

\noindent\textbf{Rotation and crops and illumination.}
The collected 4069 materials are augmented by rotation and cropping. In particular, we extract 168 unique crops for each material, at different scales. The crops are extracted in the following way. The original material is rotated 8 times ($0\degree$, $45\degree$, $90\degree$, $135\degree$, $180\degree$, $225\degree$, $270\degree$, $315\degree$)--for angles that are not multiple of $90\degree$, we first tile the material, apply the rotation, and finally crop the center of the material at 4K resolution. For each rotation, we extract 16 non-overlapping crops at 1K resolution, 4 at 2K resolution, 1 at 4K resolution, for a total of 21 crops per rotation. All crops are resized to a resolution of $1024\times1024$ pixels.
Each crop is rendered under different environment illuminations (2 outdoors, 2 indoors, and 1 studio light), for a total of 840 renders per material. For each render, a random rotation is applied to the environment map.
The resulting dataset consists of 683,592 material crops and 3,417,960 renders. Fig.~\ref{fig:render_samples} shows the five renderings under different environment illumination for 4 different materials from the dataset. \suppmat{We include the 5 environment maps used for rendering in the supplemental materials.}

\noindent\textbf{Rendering.}
We render each material displacing the mesh using the available Height map, thus introducing realistic cast shadows in the rendering. However, to maintain a pixel match with the maps, required for supervised SVBRDF estimation tasks, we opt for an orthographic rendering. This introduces yet another issue, as specular reflections are lost on perfectly flat surfaces due to rays being parallel when rendering with an orthographic camera. To restore specular reflection while keeping pixel match and cast shadows, we split the rendering in two separate passes.
Similar to the data used to train ControlMat~\cite{vecchio2023controlmat}, we first render the diffuse component of the material using an orthographic camera with displacement enabled, we then render the specular component using a perspective camera, while disabling the displacement. The two renders are then added back to get the final render. Fig.~\ref{fig:render_pass} shows the output of the two render passes and the final result. 
We evaluate each augmentation in an ablation study in the Supplemental Materials.
\input{figures/render_pass/render_pass}

\subsection{Compatibility with previous datasets}
To ensure compatibility and convenient combination between the previously existing large scale material dataset~\cite{Deschaintre18} and ours, we process it in the same way we did ours. We first extract the base color and metallic maps from the existing Diffuse and Specular maps. We extract 5 crops for each material due to the 2048x2048 resolution of the existing dataset. As the existing materials are not tileable we do not apply the rotation augmentation, and directly render each crops under 5 different lighting. This resulting in an additional $40,700$ renderings of $8,140$ material crops. We will provide these materials and renderings alongside our dataset with clear annotation of origin.

\subsection{Dataset Statistics}
We report here statistics of our dataset in terms of categories, tags, creation methodology, stationarity, and number of asset per source.

\noindent\textbf{Categories.} We report the category histogram in Fig.~\ref{fig:categories}, showing a wide diversity of appearances, and compensating for some gaps in the distribution of previously available materials.

\noindent\textbf{Tags.} We have $21,737$ tags from $1,239$ unique tags associated with the materials. For example, the top five most represented tags are floor(924), dirty(543), man (504), wood (474), brown (444) and wall(424). Each material has between 0 (371 assets didn't have associated tags) and 20 tags with an average of 5.74, a standard variation of 4.35 and a median of 5.

\noindent\textbf{Creation methodology.} Our assets are created through 4 different approaches: manual approximation (299), photometry/acquisition (651), procedural generation (2689), blends (332), unknown (98).

\noindent\textbf{Stationarity.} Our dataset contains 3061 stationary and 1008 non-stationary materials. We compute stationarity by extracting 8 random crop of the basecolor and height map and comparing the CLIP cross-similarity matrix. If more than 12.5\% of the cross-similarities are below a threshold (we empirically found 0.9 to work well), the material is not considered stationary.  

\noindent\textbf{Assets per source.} We source 1425 materials from AmbientCG~\cite{AmbientCG}, 278 materials from CGBookCase~\cite{CGBookCase}, 387 materials from PolyHeaven~\cite{PolyHeaven}, 647 materials from ShateTexture~\cite{ShareTextures}, 571 materials from TextureCan~\cite{TextureCan}, and 224 materials from the artist Julio Sillet. We add 332 blended materials and 205 variations of materials from AmbientCG~\cite{AmbientCG}.

\input{plots/categories}

%% file: figures/blend_samples/blend_samples.tex
\begin{figure}
    \centering
    \setlength{\tabcolsep}{.5pt}
    \begin{tabular}{cccc}
        Mat. 1 & Mat. 2 & Blend & Mask \\
        \vspace{-1mm}\includegraphics[width=0.245\linewidth, height=0.245\linewidth]{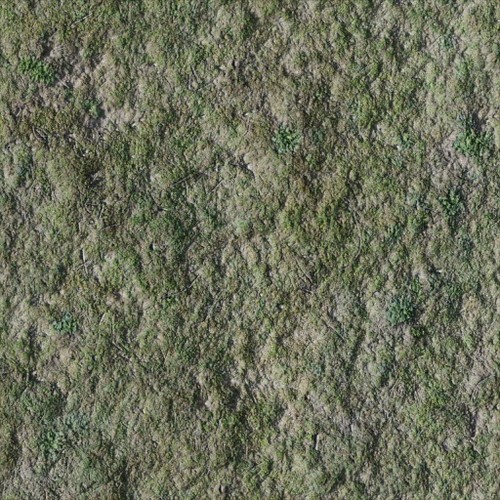} &
        \includegraphics[width=0.245\linewidth, height=0.245\linewidth]{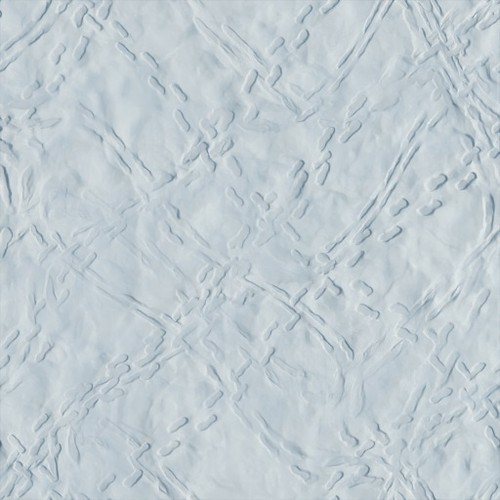} &
        \includegraphics[width=0.245\linewidth, height=0.245\linewidth]{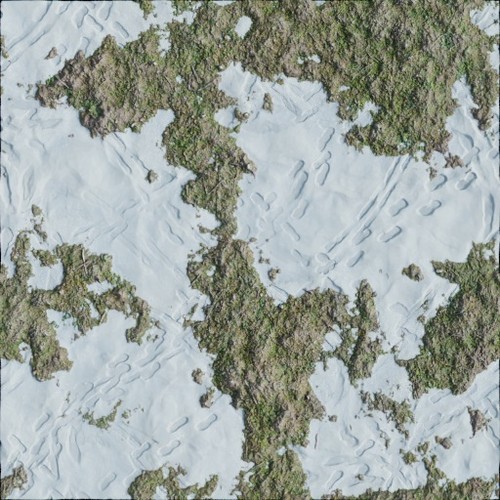} &
        \includegraphics[width=0.245\linewidth, height=0.245\linewidth]{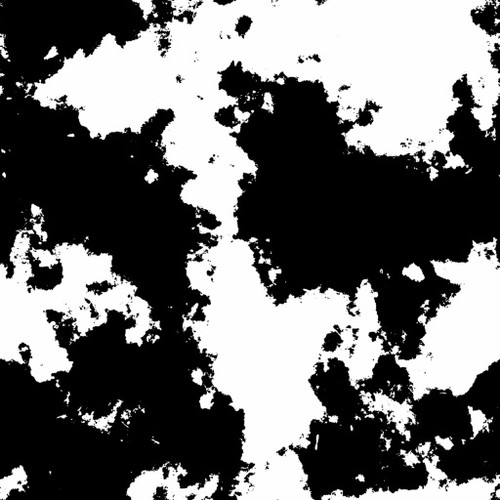} \\
    
        \vspace{-1mm}\includegraphics[width=0.245\linewidth, height=0.245\linewidth]{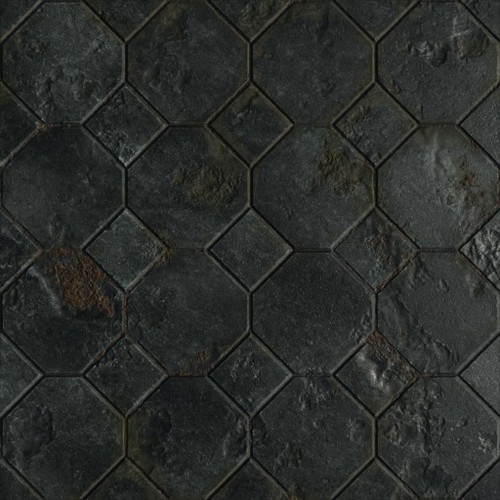} &
        \includegraphics[width=0.245\linewidth, height=0.245\linewidth]{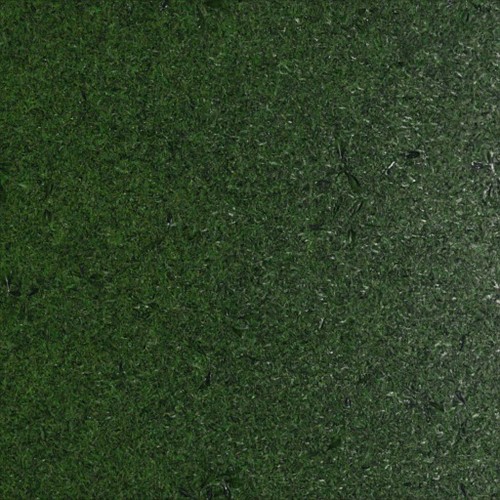} &
        \includegraphics[width=0.245\linewidth, height=0.245\linewidth]{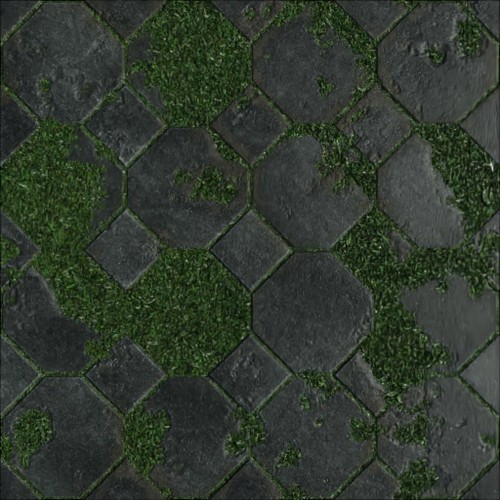} &
        \includegraphics[width=0.245\linewidth, height=0.245\linewidth]{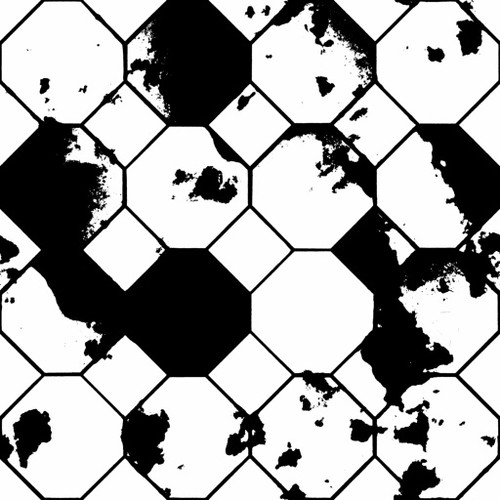} \\
    \end{tabular}
    
    \caption{\textbf{Samples of height based material blends.} We show renderings of both materials being blended, the result of our blending, and the height-based computed mask.}
    \label{fig:blend_samples}
\end{figure}

%% file: figures/render_samples/render_samples.tex
\begin{figure}
    \centering
    \setlength{\tabcolsep}{.5pt}
    \begin{tabular}{ccccc}
        Studio & Indoor 1 & Indoor 2 & Outdoor 1 & Outdoor 2 \\
        \vspace{-1mm}\includegraphics[width=0.195\linewidth]{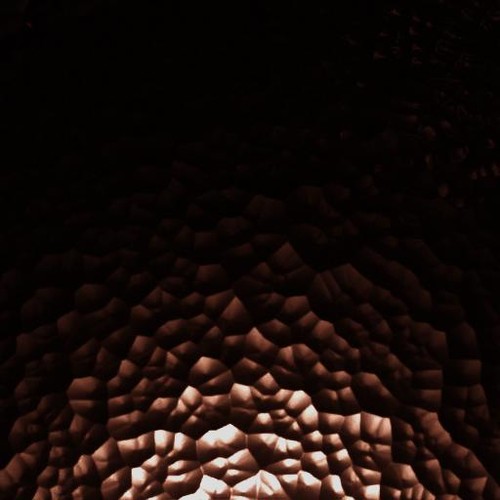} &
        \includegraphics[width=0.195\linewidth]{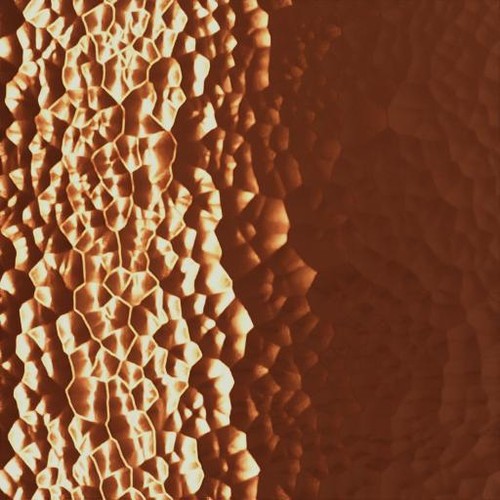} &
        \includegraphics[width=0.195\linewidth]{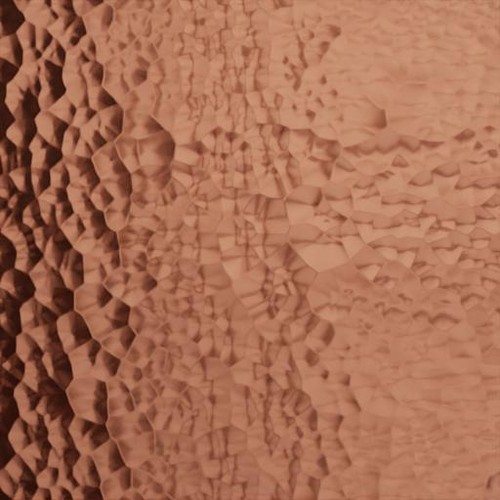} &
        \includegraphics[width=0.195\linewidth]{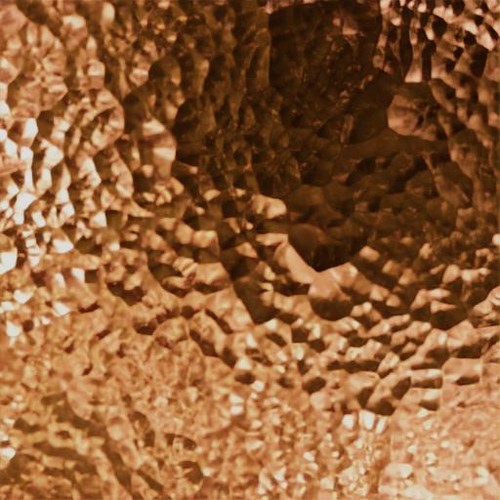} &
        \includegraphics[width=0.195\linewidth]{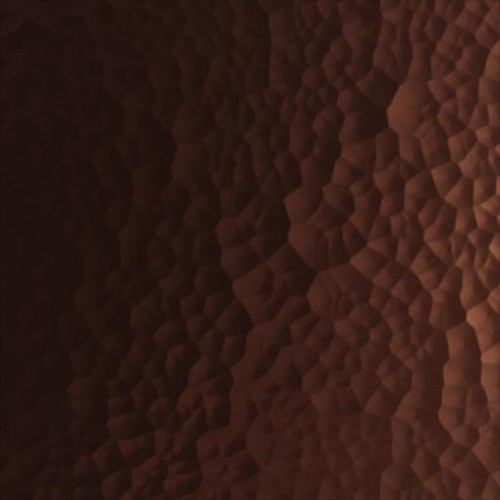} \\
    
        \vspace{-1mm}\includegraphics[width=0.195\linewidth]{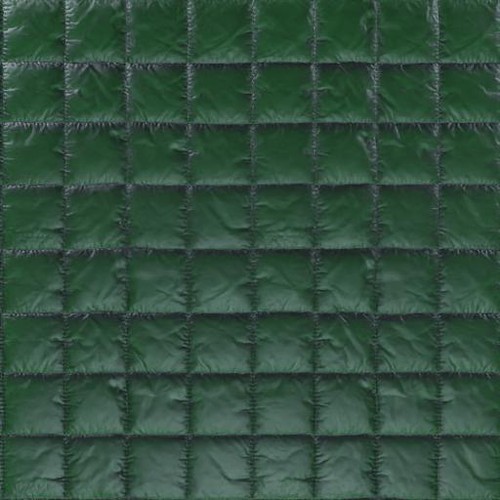} &
        \includegraphics[width=0.195\linewidth]{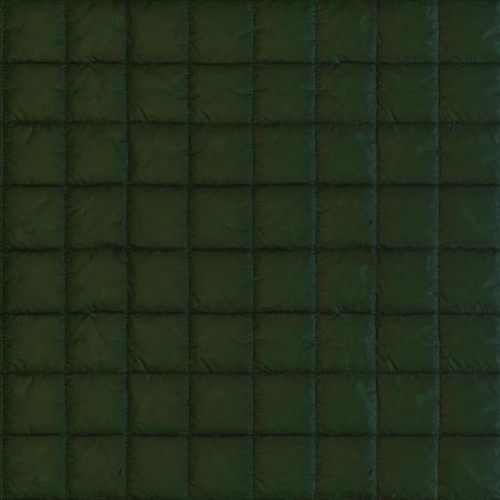} &
        \includegraphics[width=0.195\linewidth]{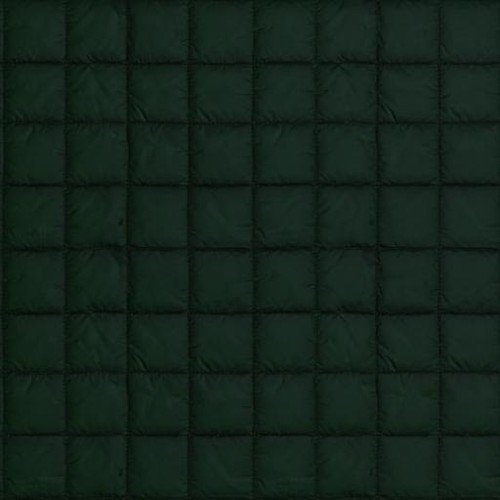} &
        \includegraphics[width=0.195\linewidth]{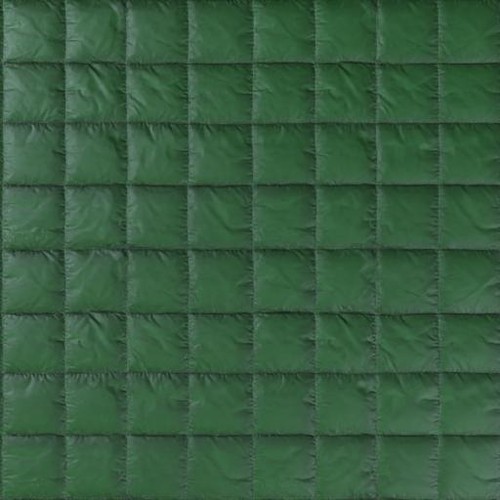} &
        \includegraphics[width=0.195\linewidth]{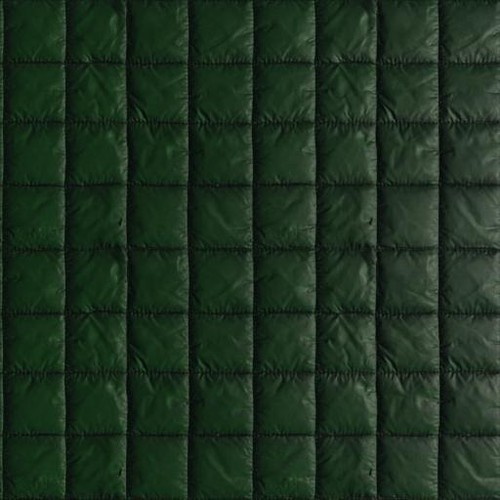} \\
    
        \vspace{-1mm}\includegraphics[width=0.195\linewidth]{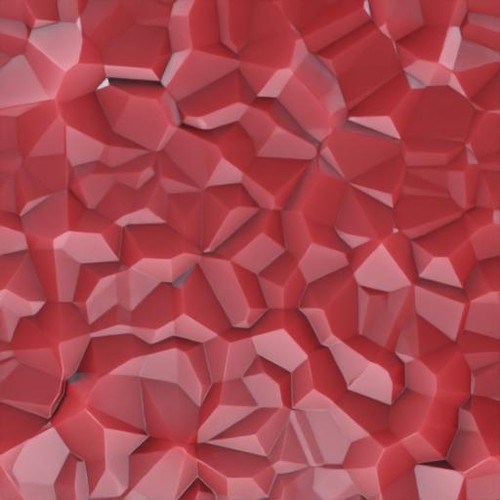} &
        \includegraphics[width=0.195\linewidth]{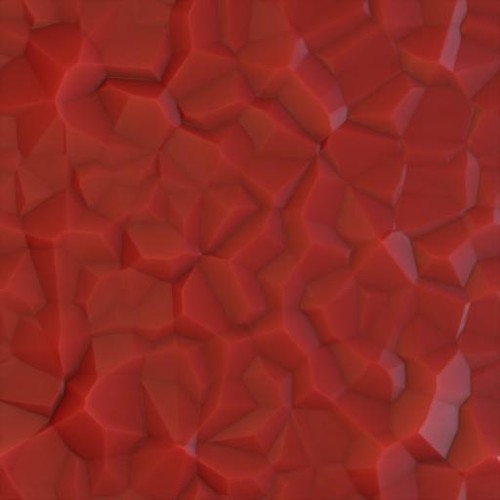} &
        \includegraphics[width=0.195\linewidth]{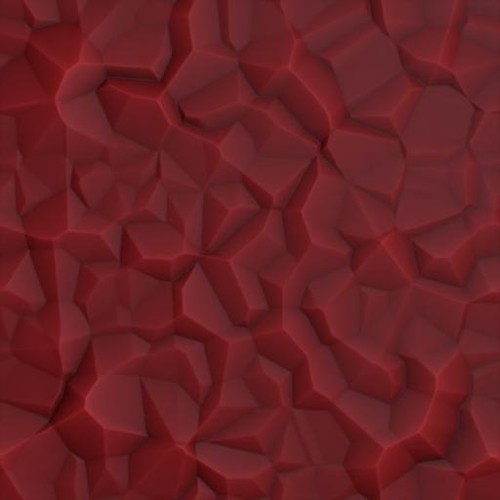} &
        \includegraphics[width=0.195\linewidth]{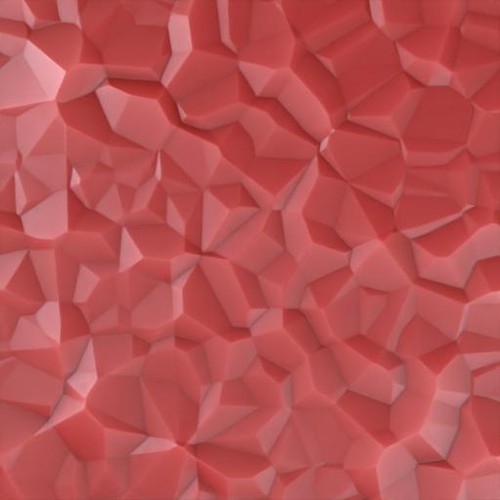} &
        \includegraphics[width=0.195\linewidth]{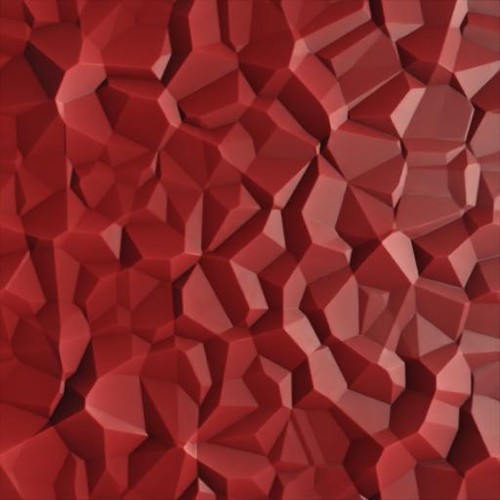} \\

        \vspace{-1mm}\includegraphics[width=0.195\linewidth]{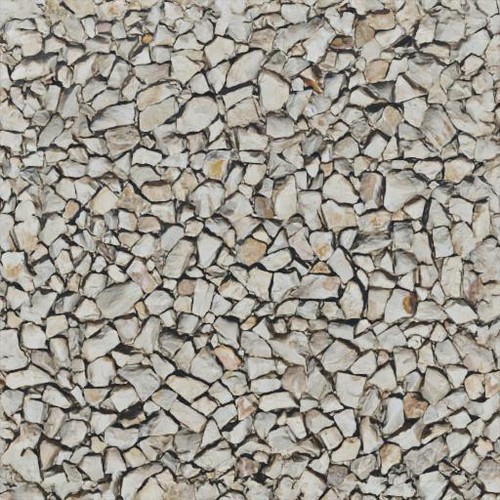} &
        \includegraphics[width=0.195\linewidth]{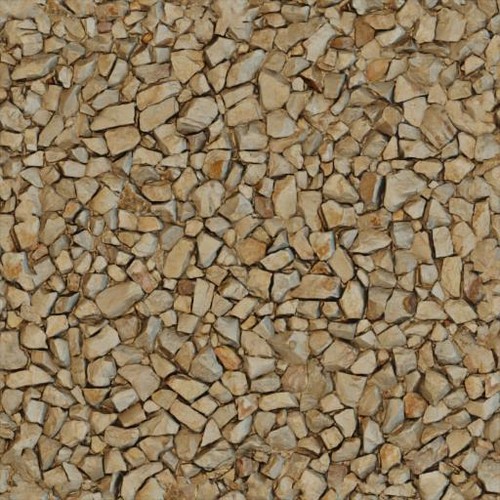} &
        \includegraphics[width=0.195\linewidth]{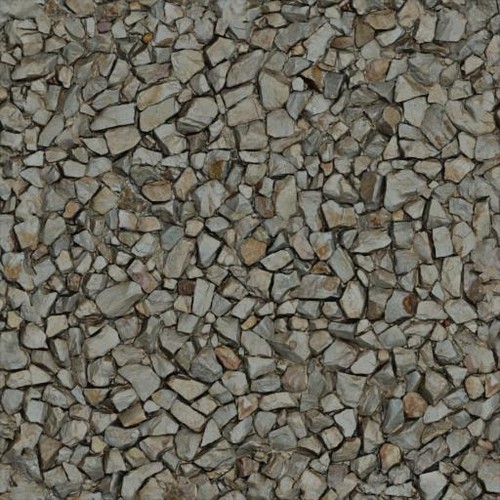} &
        \includegraphics[width=0.195\linewidth]{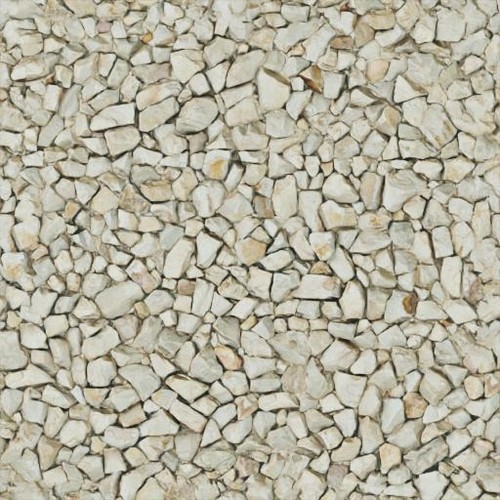} &
        \includegraphics[width=0.195\linewidth]{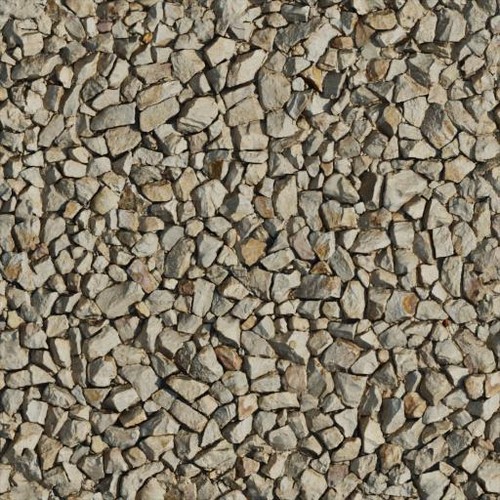} \\
    \end{tabular}
    
    \caption{\textbf{Renderings under various environment maps.} We show four materials (Metal, Leather, Plastic and Pebbles) from the dataset rendered under the 5 chosen environment maps.}
    \label{fig:render_samples}
\end{figure}

%% file: figures/render_pass/render_pass.tex
\begin{figure}
    \centering
    \setlength{\tabcolsep}{.5pt}
    \begin{tabular}{ccc}
        Diffuse Pass & Glossy Pass & Full Render \\
        \vspace{-1mm}\includegraphics[width=0.33\linewidth]{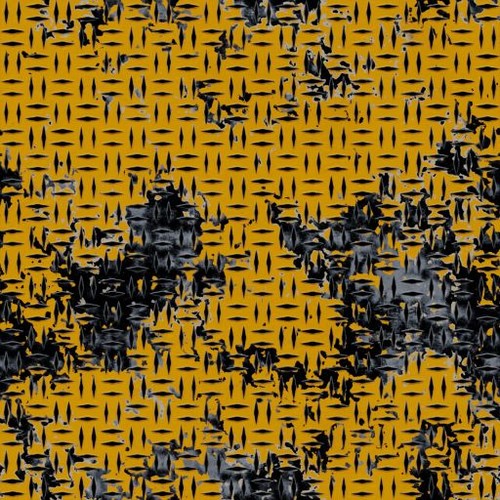} &
        \includegraphics[width=0.33\linewidth]{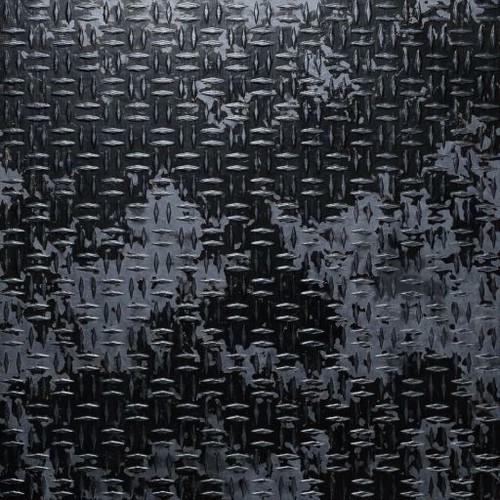} &
        \includegraphics[width=0.33\linewidth]{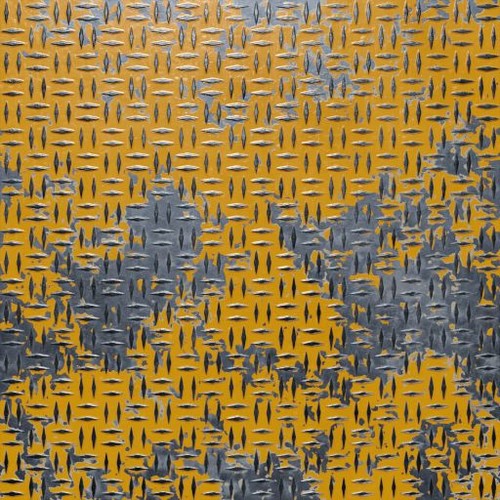} \\
    
        \vspace{-1mm}\includegraphics[width=0.33\linewidth]{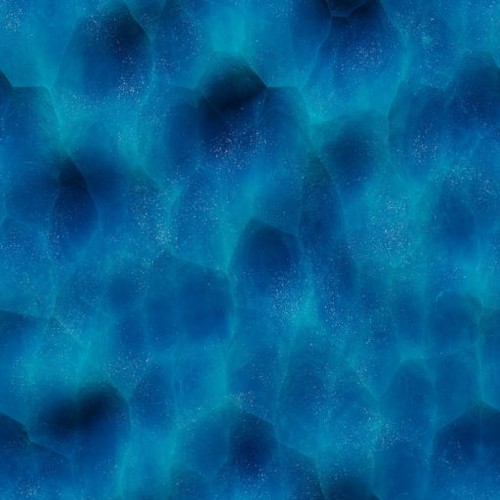} &
        \includegraphics[width=0.33\linewidth]{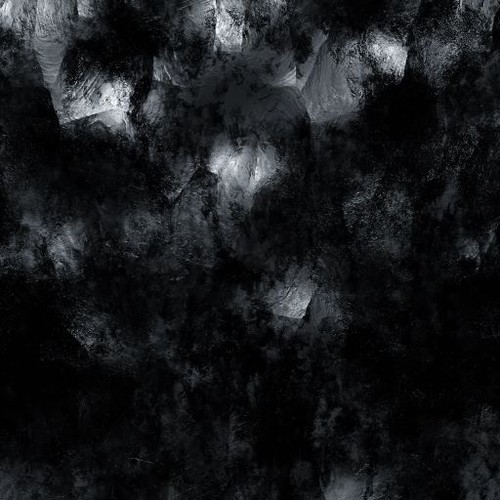} &
        \includegraphics[width=0.33\linewidth]{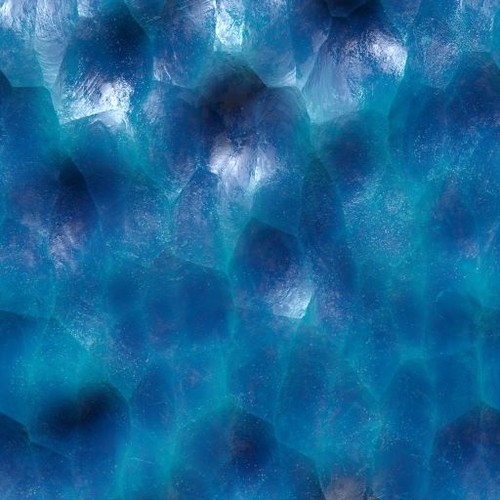} \\

        \vspace{-1mm}\includegraphics[width=0.33\linewidth]{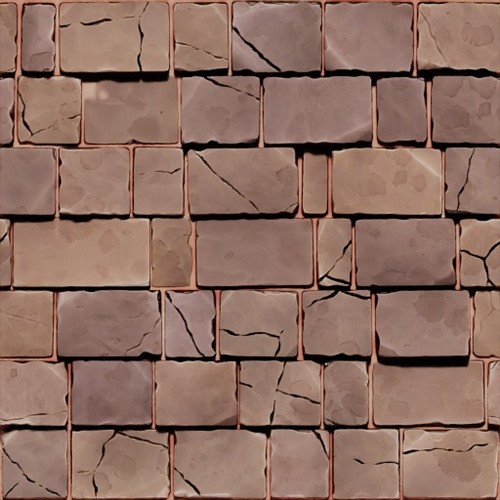} &
        \includegraphics[width=0.33\linewidth]{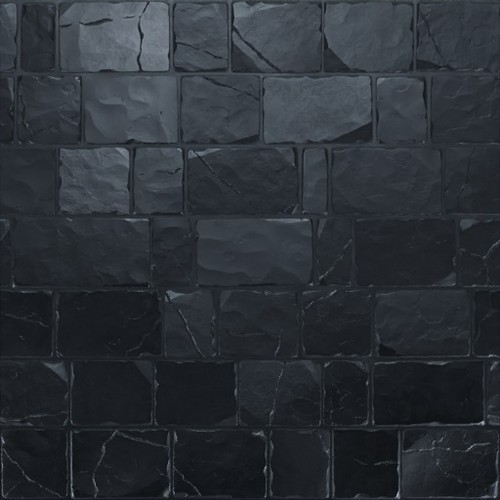} &
        \includegraphics[width=0.33\linewidth]{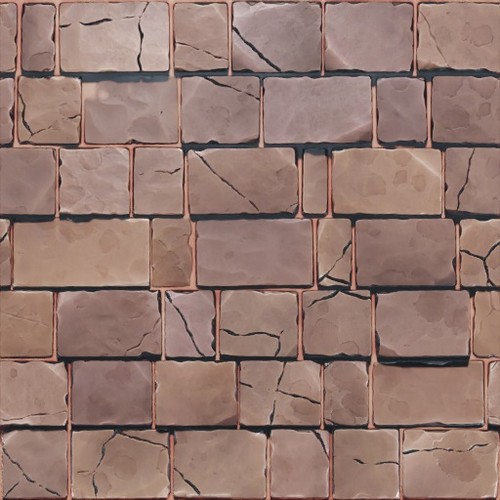} \\
    \end{tabular}
    
    \caption{\textbf{Render samples using the two-pass strategy.} This ensures that the maps and the rendering are well aligned, avoiding parrallax effects, but preserving specular highlights.}
    \label{fig:render_pass}
\end{figure}

%% file: plots/categories.tex
\pgfplotstableread[row sep=\\,col sep=&]{
    interval          & MatSynth & Desch20 \\
    Blends            & 332 & 0     \\
    Ceramic           & 431 & 320   \\
    Concrete          & 226 & 0     \\
    Fabric            & 390 & 63    \\
    Ground            & 414 & 109   \\
    Leather           & 144 & 121   \\
    Marble            & 136 & 0     \\
    Metal             & 349 & 333   \\
    Misc              & 257 & 0     \\
    Plaster           & 144 & 74    \\
    Plastic           & 153 & 57    \\
    Stone             & 308 & 292   \\
    Terracotta        & 225 &       \\
    Wood              & 560 & 259   \\
}\categories

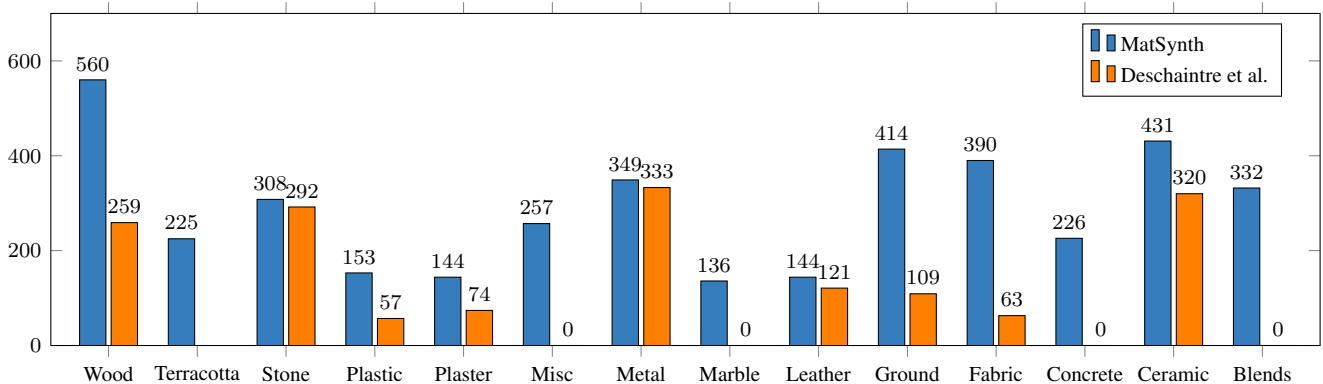
\begin{figure*}
    \centering
    \begin{tikzpicture}[every node/.style={font=\footnotesize}]
        \begin{axis}[
                height=6cm,
                ybar,
                nodes near coords, 
                bar width=0.35cm,
                x=1.18cm,
                enlarge x limits=.05,
                symbolic x coords={Wood, Terracotta, Stone, Plastic, Plaster, Misc, Metal, Marble, Leather, Ground, Fabric, Concrete, Ceramic, Blends},
                ymin=0,
                ymax=700,
                legend style={
                    legend pos=north east,
                    cells={anchor=west},
                    font=\footnotesize,
                }
            ]
            \addplot[fill=cvprblue] table[y=MatSynth,x=interval]{\categories};
            \addplot[fill=orange] table[y=Desch20,x=interval]{\categories};
            \legend{MatSynth, Deschaintre et al.}
        \end{axis}
    \end{tikzpicture}
    \caption{\textbf{Categories distribution of \nameMethod and \citet{Deschaintre18}'s dataset.} As there are no duplicates in the two datasets, the two datasets can be joined to combine their benefits. We show here that our dataset significantly enriches the previously available data, reducing the inequalities in the category distributions of the previous dataset, in particular for Fabric of Ground categories.}
    \label{fig:categories}
\end{figure*}

%% file: sec/5_results.tex
\input{tables/estimation_results}

\section{Experimental Results}
\label{sec:results}

In this section, we first introduce the dataset evaluation methodology and test set, we then briefly present the different methods used for evaluation and their training strategy. Finally, we show quantitative and qualitative results.

\subsection{Test dataset}
We build a static test set, with the aim of simplifying the comparison of future work, by handpicking 5 unique samples for each category. We further ensure that none of the chosen samples is a simple variation (e.g.: color or scale) of other items in the training set to avoid any bias. We merge this dataset, consisting of 65 materials, with the test set from Deschaintre et al.~\cite{Deschaintre20}, for a total of 89 test materials.

\subsection{Evaluation Methodology}
To evaluate our dataset we propose a performance comparison between different state of the art methods when trained on the on the dataset by Deschaintre et al.~\cite{Deschaintre18}, and on \nameMethod. 

Evaluation is carried on two main materials--related tasks: estimation and generation. For both tasks, we look at methods with publicly available implementations. In particular, we evaluate Deschaintre et al.~\cite{Deschaintre18} and SurfaceNet~\cite{vecchio2021surfacenet} for materials capture performance, and MatFuse~\cite{vecchio2023matfuse} for material generation. In both cases, we focus on the diffuse/specular rendering workflow for an easier comparison with the current state of the art. \suppmat{We additionally propose an evaluation of SurfaceNet trained using the basecolor/metallic workflow in Supplemental material, to set a new baseline for future work shifting to this workflow.}
\input{figures/estim_synth}

\input{figures/estim_real}

\subsection{Training Strategy}
All methods are trained on PC with a single NVIDIA RTX 4090 GPU. Training of \citet{Deschaintre18} and SurfaceNet~\cite{vecchio2021surfacenet} is carried out on a flash-lit version of the dataset, as these methods are designed for SVBRDF estimation from flash-lit pictures.

\noindent\textbf{Deschaintre et al. (2018)}
We train the method, using the author's publicly available Tensorflow implementation, on a combination of our dataset and the original one. Following the original paper, we train the model for $400,000$ steps (72 hours) with a batch size of 8 and a learning rate of $2^{-5}$.

\noindent\textbf{SurfaceNet}
Similarly to the training of \citet{Deschaintre18}, we use the author's original Pytorch implementation and the combined dataset. We also follow the original paper and train the model for $400,000$ steps (96 hours) with a batch size of 8 and a learning rate of $2^{-5}$. We include the adversarial discriminator after $100,000$ steps.

\noindent\textbf{Diffusion Model}
We evaluate the quality improvement of generation provided by our dataset on the recent MatFuse~\cite{vecchio2023matfuse}. We evaluate the performance both when fine-tuning the diffusion model and when training from scratch. For the first experiment, we use the same multi-encoders compression model, pre-trained on the \citet{Deschaintre18}'s dataset, but fine-tune the latent diffusion model itself. In particular, the diffusion model is fine-tuned for 150,000 iterations with a batch size of $20$ and a learning rate of $10^{-5}$. We use a linear learning rate warm-up until step 10,000. \\
For the second experiment, we train the entire model, on our dataset. In particular, we train the autoencoder model for 1.000.000 iterations with a batch size of $4$ and a learning rate of $10^{-4}$. We include the adversarial discriminator after $300,000$ steps. Finally, we train the diffusion model for $500,000$ iterations with a batch size of $16$ using an AdamW~\cite{adamw} optimizer, with a learning rate value of $10^{-4}$, with a linear learning rate warm-up until step 10,000. \\
For inference, following the original paper, we denoise using a DDIM sampling schedule~\cite{song2020denoising} with $50$ timesteps. 

\subsection{SVBRDF Estimation Results}
\input{figures/generative_results}

We evaluate the quality improvement using our dataset for two method: \citet{Deschaintre18} and SurfaceNet\cite{vecchio2021surfacenet}. 

We retrain both of these methods on our dataset and report the RMSE for each material map (cosine distance for normals), as well as average RMSE and more perceptual metrics (LPIPS and SSIM) between renderings under 5 different environment illumination. We report the results in Table~\ref{tab:comparison_quantitative_rmse} and Table~\ref{tab:comparison_quantitative_perceptual}, showing improvement accross the board, in particular for more challenging maps such as Roughness and Specularity.

We further show a qualitative comparison on synthetic data in Figure~\ref{fig:comparison_acquisition_synth}, as well as real real data captured with a phone camera in Figure~\ref{fig:comparison_acquisition_real}, demonstrating that the additional data helps achieve better material capture quality, in particular for Surface Net~\cite{vecchio2021surfacenet} which can leverage the new data best through its discriminator.

\subsection{Materials Generation Results}
We evaluate material generation using MatFuse~\cite{vecchio2023matfuse}, trained on \citet{Deschaintre18}'s dataset only, and comparing it against a version fine-tuned with \nameMethod and one fully trained on it. As material maps have different distributions than natural images, we report FID on renderings under environment lighting. For fairness, we only use \citet{Deschaintre18}'s dataset to compute the Ground-Truth distribution and compute the FID for each method using $1,000$ random samples. Before fine-tuning we find a FID score (lower is better) of $239.9$, $210.3$ after fine-tuning, and $89.84$ when training from scratch. 
We believe the performance difference to be caused by the retraining of the autoencoder using MatSynth, while the other experiments used the original VQ-VAE from MatFuse, trained on \citet{Deschaintre18}'s dataset only.

We further evaluate the benefit of the additional data for diversity and quality. We show in Table~\ref{tab:generated_classes_distribution} the percentage of samples belonging to different classes, and show a qualitative comparison in Fig.~\ref{fig:generative_comparison}. Both evaluations show a wider diversity of results, and Fig.~\ref{fig:generative_comparison} shows that the generation results are more realistic after fine-tuning or when training on \nameMethod from scratch.

\begin{table}[]
    \centering
    \begin{tabular}{c|ccc}
        \toprule
         & original & fine-tuned & full training \\
        \midrule
        Wood & 36.1\% & 25.72\% & 7.5\% \\
        Ceramic & 1.1\% & 12.51\% & 8.5\% \\
        Stone & 21.5\% & 14.8\% & 7.3\% \\
        Metal & 3.45\% & 8.4\% & 6.4\% \\
        Fabric & 0.33\% & 7.34\% & 7.8\% \\
        Ground & 1.8\% & 16.2\% & 6.4\% \\
        Others & 35.72\% & 15.03\% & 56.1\% \\
        \bottomrule
    \end{tabular}
    \caption{\textbf{Generated materials class distribution.} As expected, we observe that after fine-tuning the generative model create more diverse materials. We classify our results using CLIP~\cite{clip} based zero-shot classification.}
    \label{tab:generated_classes_distribution}
\end{table}

%% file: tables/estimation_results.tex
\begin{table*}
    \centering
    \begin{tabular}{lrrrr}
        \toprule
        \textbf{Image Type} & Deschaintre (\cite{Deschaintre18}) & Deschaintre (\textbf{MatSynth}) & SurfaceNet (\cite{vecchio2021surfacenet}) & SurfaceNet (\textbf{MatSynth}) \\
        \hline
        Renderings & 0.172 & 0.160 & 0.161 & \textbf{0.135} \\ 
        Diffuse & 0.100 & \textbf{0.093} & 0.119 & 0.094 \\ 
        Normal \small{(Cos dist)} & 0.576 & 0.573 & 0.600 & \textbf{0.544} \\ 
        Roughness & 0.322 & 0.215 & 0.221 & \textbf{0.162} \\ 
        Specular & 0.119 & 0.080 & 0.086 & \textbf{0.057} \\ 
        \bottomrule
    \end{tabular}
    \caption{\textbf{Quantitative results.} We report the RMSE$\downarrow$ between predicted and ground-truth maps and renderings (averaged over renderings under 5 environment lightings), except for Normal maps for which we report the cosine error$\downarrow$. We see that the additional data provided by \nameMethod enables higher quality material acquisition both in terms of parameters and renderings comparisons.}
    \label{tab:comparison_quantitative_rmse}
\end{table*}

\begin{table*}
    \centering
    \begin{tabular}{llrrrr}
        \toprule
        \textbf{Image Type} & \textbf{Metric} & Deschaintre (\cite{Deschaintre18}) & Deschaintre (\textbf{MatSynth}) & SurfaceNet (\cite{vecchio2021surfacenet}) & SurfaceNet (\textbf{MatSynth}) \\
        \midrule
        \multirow{2}{*}{Renderings} & SSIM & 0.532 & 0.560 & 0.494 & \textbf{0.613} \\ 
                                    & LPIPS & 0.560 & 0.294 & 0.395 & \textbf{0.281} \\
        \bottomrule
    \end{tabular}
    \caption{\textbf{Perceptual quantitative results.} We report here the perceptual metrics for the Renderings, rendered under 5 environment lightings (which can be intepreted as natural images) SSIM$\uparrow$ and LPIPS$\downarrow$. Similar to the RMSE evaluation, additional data significantly improves the rendering matching's quality.}
    \label{tab:comparison_quantitative_perceptual}
\end{table*}

%% file: figures/estim_synth.tex
\begin{figure*}
    \setlength{\tabcolsep}{.5pt}
    \hspace{-0.15cm}
    \begin{tabular} {ccccccccccccc} 
        \vspace{0.25cm} \\ %
        \hspace{-1mm} & Input & Diffuse & Normal & Rough. & Specular & Render & Input & Diffuse & Normal & Rough. & Specular & Render \\
        
        \vspace{-1mm} \hspace{-1mm} \begin{sideways} \hspace{5mm}\scriptsize{GT} \end{sideways}\hspace{1mm} & & 
        \includegraphics[width=0.08\linewidth]{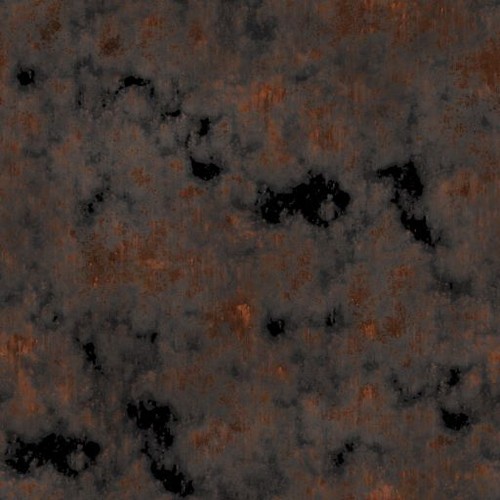} & 
        \includegraphics[width=0.08\linewidth]{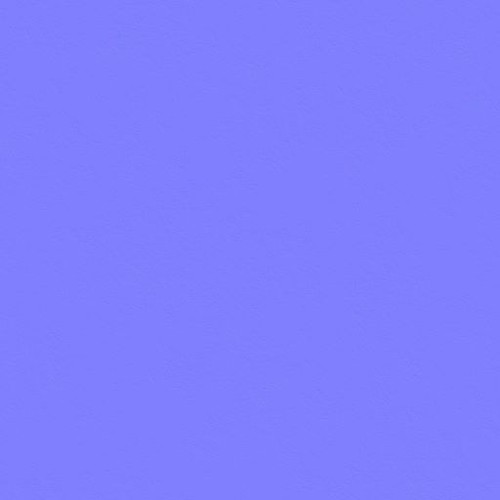} & 
        \includegraphics[width=0.08\linewidth]{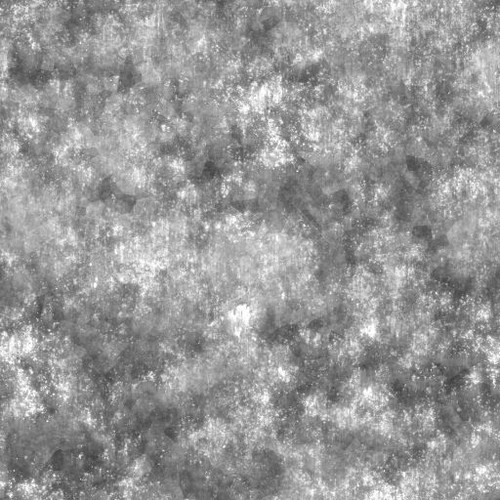} & 
        \includegraphics[width=0.08\linewidth]{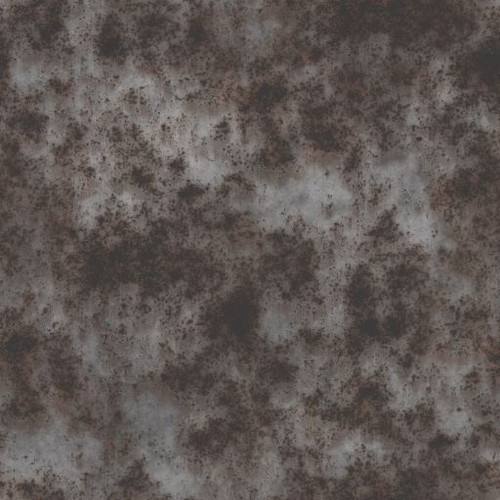} & 
        \includegraphics[width=0.08\linewidth]{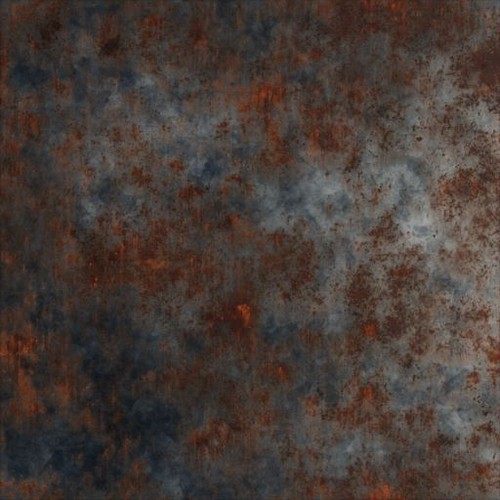} &
        \hspace{0.25mm} & 
        \includegraphics[width=0.08\linewidth]{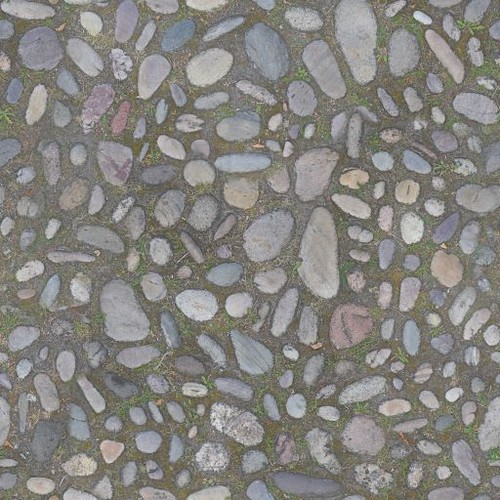} & 
        \includegraphics[width=0.08\linewidth]{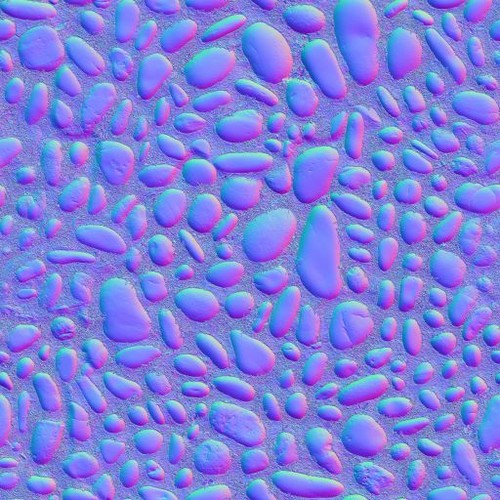} & 
        \includegraphics[width=0.08\linewidth]{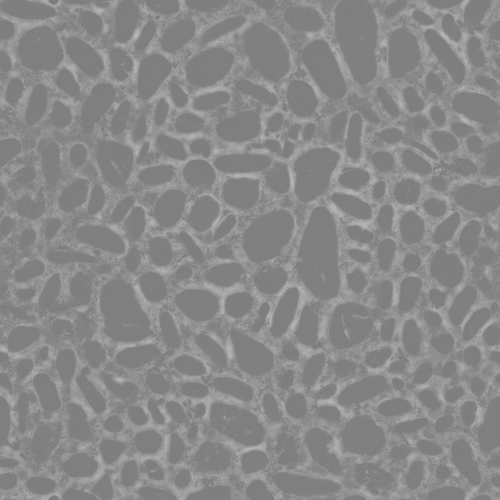} & 
        \includegraphics[width=0.08\linewidth]{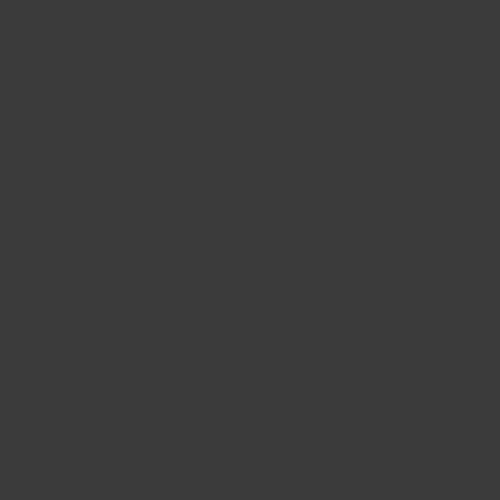} & 
        \includegraphics[width=0.08\linewidth]{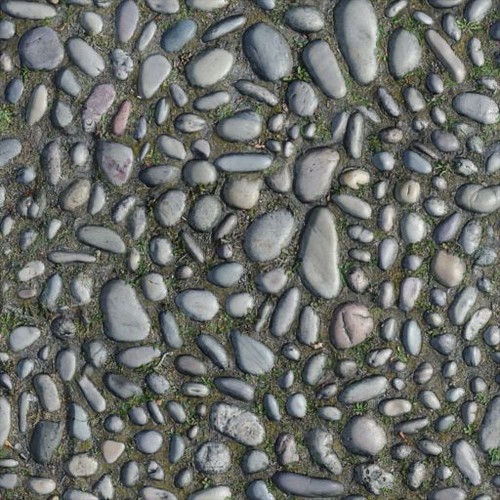} \\

        \vspace{-1mm} \hspace{-1mm} \begin{sideways} \scriptsize{Deschaintre} \end{sideways}\hspace{1mm} & 
        \includegraphics[width=0.08\linewidth]{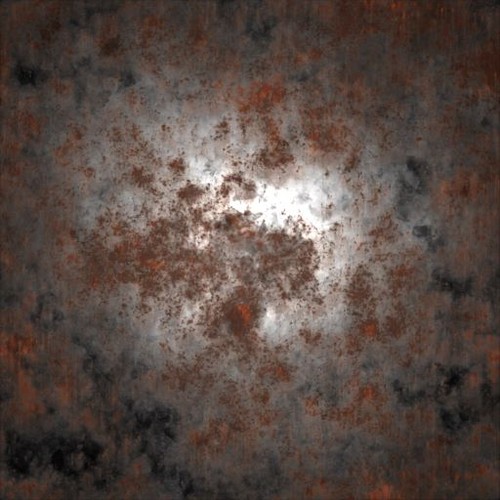} & 
        \includegraphics[width=0.08\linewidth]{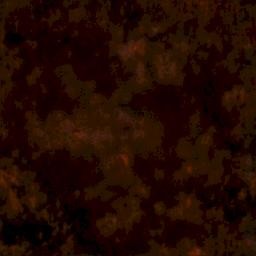} & 
        \includegraphics[width=0.08\linewidth]{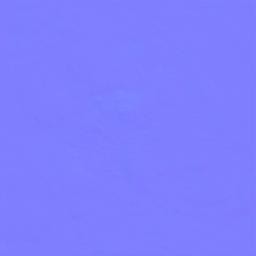} & 
        \includegraphics[width=0.08\linewidth]{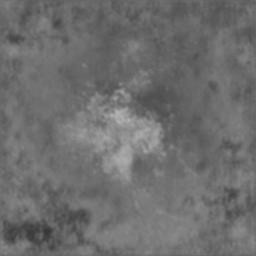} & 
        \includegraphics[width=0.08\linewidth]{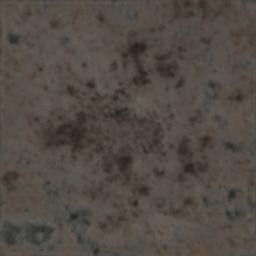} & 
        \includegraphics[width=0.08\linewidth]{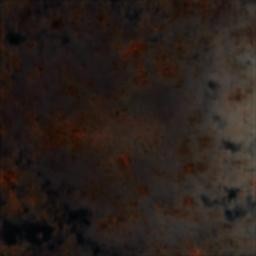} & 
        \hspace{0.25mm} \includegraphics[width=0.08\linewidth]{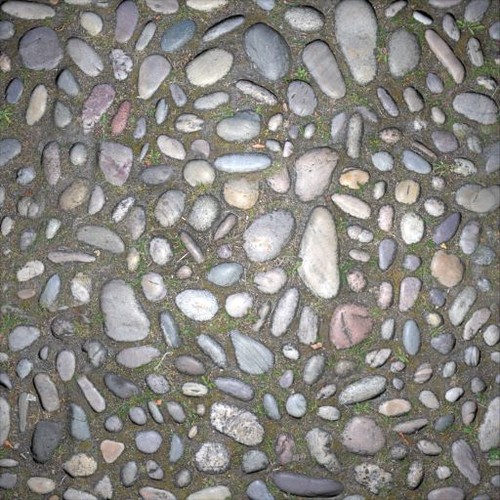} & 
        \includegraphics[width=0.08\linewidth]{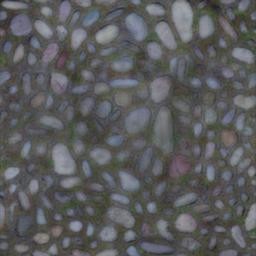} & 
        \includegraphics[width=0.08\linewidth]{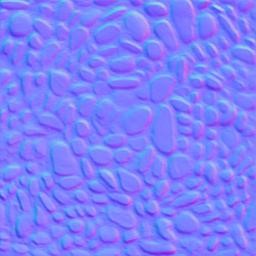} & 
        \includegraphics[width=0.08\linewidth]{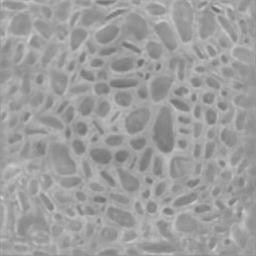} & 
        \includegraphics[width=0.08\linewidth]{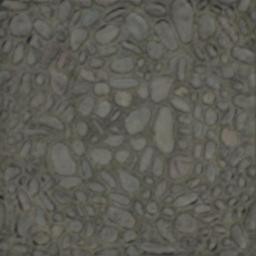} & 
        \includegraphics[width=0.08\linewidth]{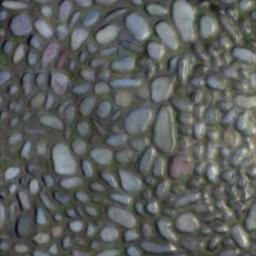} \\

        \vspace{-1mm} \hspace{-1mm} \begin{sideways} \scriptsize{Deschaintre*} \end{sideways}\hspace{1mm} & 
        \includegraphics[width=0.08\linewidth]{figures/estim_synth/36_input.jpg} & 
        \includegraphics[width=0.08\linewidth]{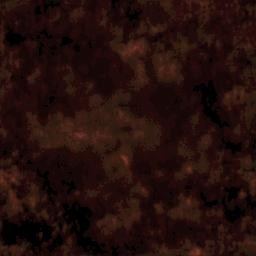} & 
        \includegraphics[width=0.08\linewidth]{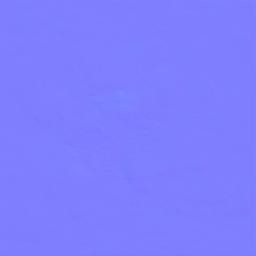} & 
        \includegraphics[width=0.08\linewidth]{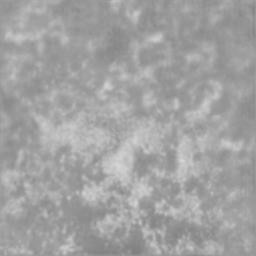} & 
        \includegraphics[width=0.08\linewidth]{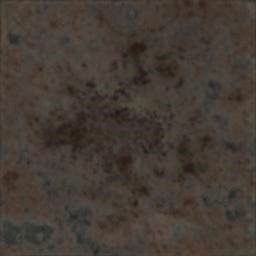} & 
        \includegraphics[width=0.08\linewidth]{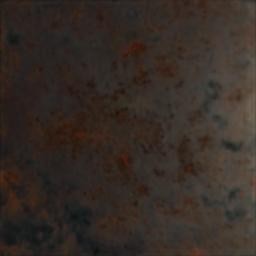} & 
        \hspace{0.25mm} \hspace{-1mm} \includegraphics[width=0.08\linewidth]{figures/estim_synth/40_input.jpg} & 
        \includegraphics[width=0.08\linewidth]{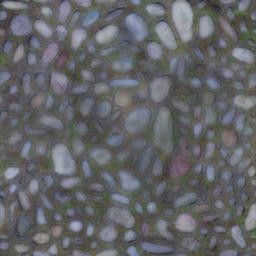} & 
        \includegraphics[width=0.08\linewidth]{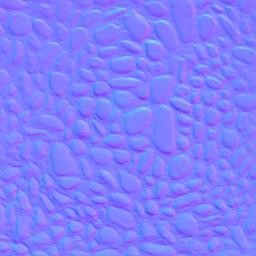} & 
        \includegraphics[width=0.08\linewidth]{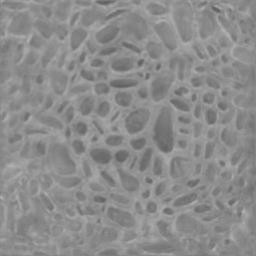} & 
        \includegraphics[width=0.08\linewidth]{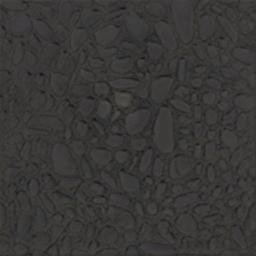} & 
        \includegraphics[width=0.08\linewidth]{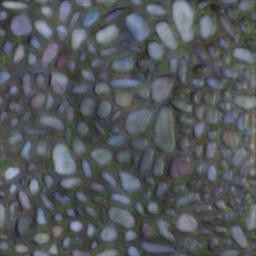} \\

        \vspace{-1mm} \hspace{-1mm} \begin{sideways} \scriptsize{SurfaceNet} \end{sideways}\hspace{1mm} & 
        \includegraphics[width=0.08\linewidth]{figures/estim_synth/36_input.jpg} & 
        \includegraphics[width=0.08\linewidth]{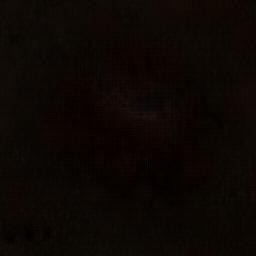} & 
        \includegraphics[width=0.08\linewidth]{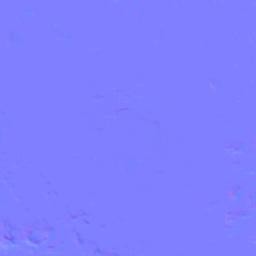} & 
        \includegraphics[width=0.08\linewidth]{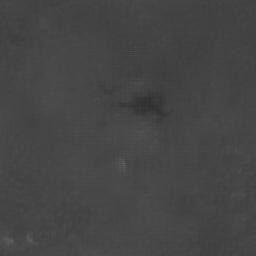} & 
        \includegraphics[width=0.08\linewidth]{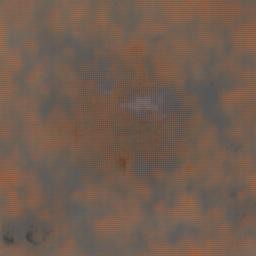} & 
        \includegraphics[width=0.08\linewidth]{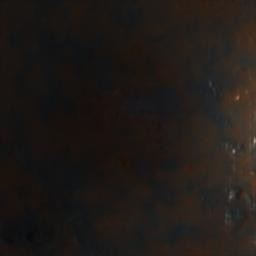} & 
        \hspace{0.25mm} \includegraphics[width=0.08\linewidth]{figures/estim_synth/40_input.jpg} & 
        \includegraphics[width=0.08\linewidth]{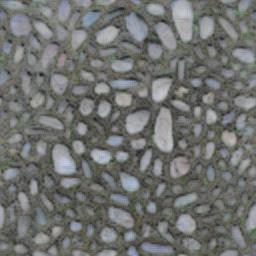} & 
        \includegraphics[width=0.08\linewidth]{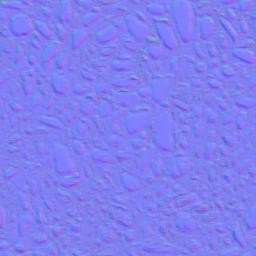} & 
        \includegraphics[width=0.08\linewidth]{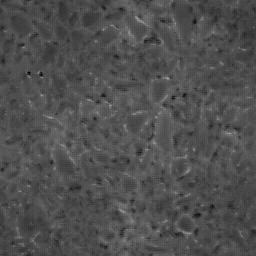} & 
        \includegraphics[width=0.08\linewidth]{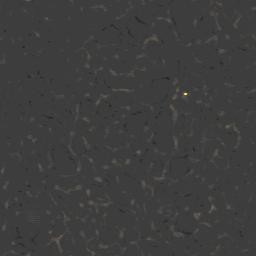} & 
        \includegraphics[width=0.08\linewidth]{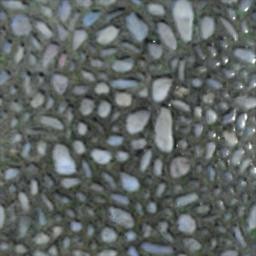} \\

        \vspace{-1mm} \hspace{-1mm} \begin{sideways} \scriptsize{SurfaceNet*} \end{sideways}\hspace{1mm} & 
        \includegraphics[width=0.08\linewidth]{figures/estim_synth/36_input.jpg} & 
        \includegraphics[width=0.08\linewidth]{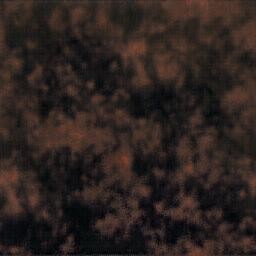} & 
        \includegraphics[width=0.08\linewidth]{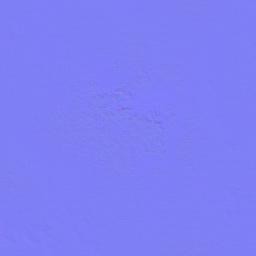} & 
        \includegraphics[width=0.08\linewidth]{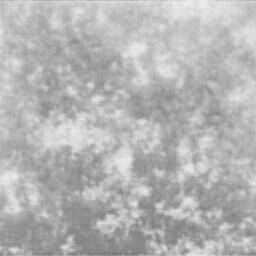} & 
        \includegraphics[width=0.08\linewidth]{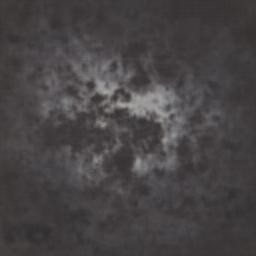} & 
        \includegraphics[width=0.08\linewidth]{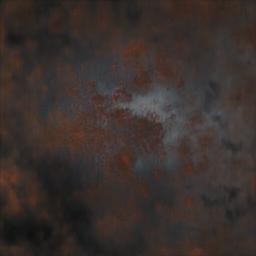} & 
        \hspace{0.25mm} \includegraphics[width=0.08\linewidth]{figures/estim_synth/40_input.jpg} & 
        \includegraphics[width=0.08\linewidth]{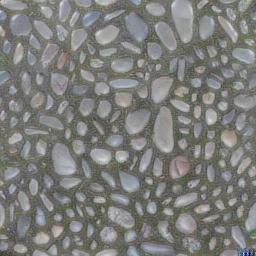} & 
        \includegraphics[width=0.08\linewidth]{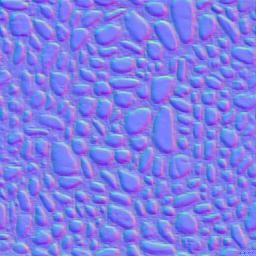} & 
        \includegraphics[width=0.08\linewidth]{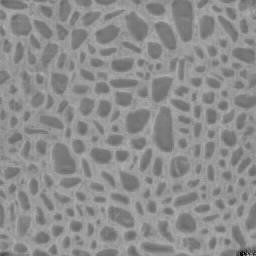} & 
        \includegraphics[width=0.08\linewidth]{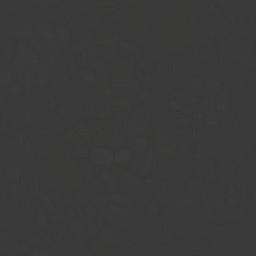} & 
        \includegraphics[width=0.08\linewidth]{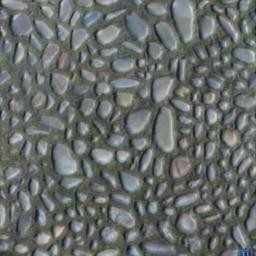} \\
        
    \end{tabular}
    \caption{\textbf{Qualitative material acquisition comparison on synthetic data.} We compare \citet{Deschaintre18} and SurfaceNet~\cite{vecchio2021surfacenet} trained only on \citet{Deschaintre18}'s dataset against the same methods trained on \nameMethod (marked with *). We can see that the fine-tuned versions better match the Ground truth, in particular for SurfaceNet in the Normal and Roughness maps.}
    \label{fig:comparison_acquisition_synth}
\end{figure*}

%% file: figures/estim_real.tex
\begin{figure*}[!ht]
    \setlength{\tabcolsep}{.5pt}
    \hspace{-0.15cm}
    \begin{tabular} {ccccccccccccc} 
        \vspace{0.25cm} \\ %
        \hspace{-1mm} & Input & Diffuse & Normal & Rough. & Specular & Render & \hspace{0.5mm} Input & Diffuse & Normal & Rough. & Specular & Render \\

        \vspace{-1mm} \begin{sideways} \scriptsize{Deschaintre} \end{sideways}\hspace{0.5mm} & 
        \includegraphics[width=0.08\linewidth]{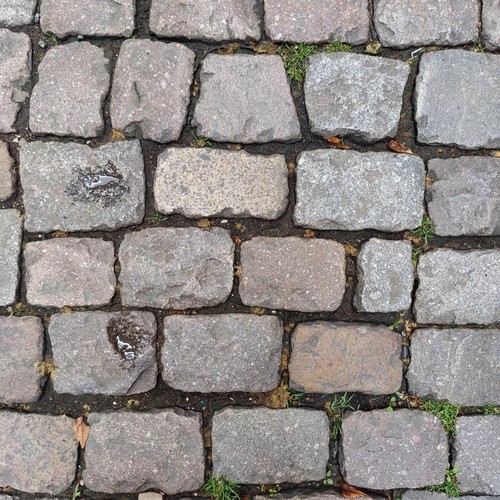} & 
        \includegraphics[width=0.08\linewidth]{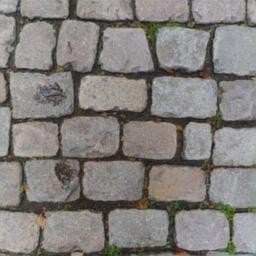} & 
        \includegraphics[width=0.08\linewidth]{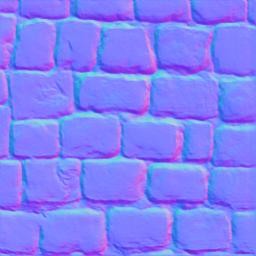} & 
        \includegraphics[width=0.08\linewidth]{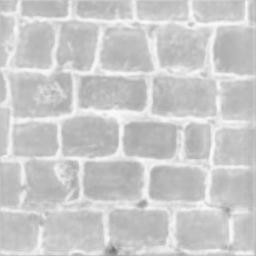} & 
        \includegraphics[width=0.08\linewidth]{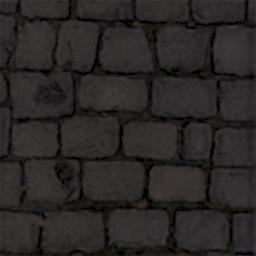} & 
        \includegraphics[width=0.08\linewidth]{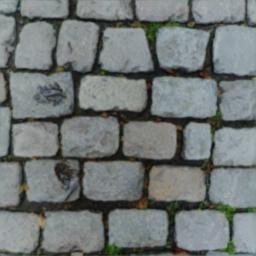} &  
        \hspace{0.25mm} \includegraphics[width=0.08\linewidth]{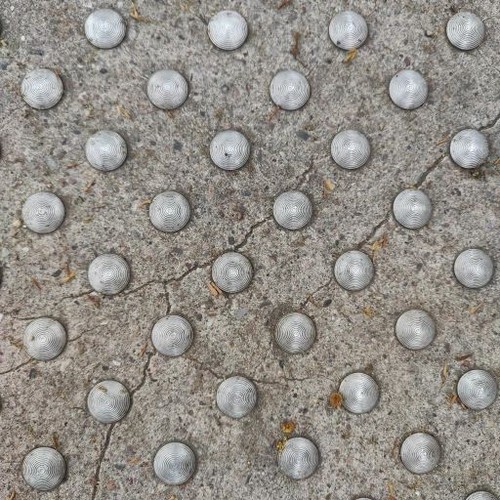} & 
        \includegraphics[width=0.08\linewidth]{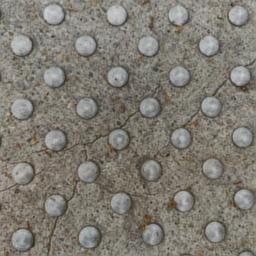} & 
        \includegraphics[width=0.08\linewidth]{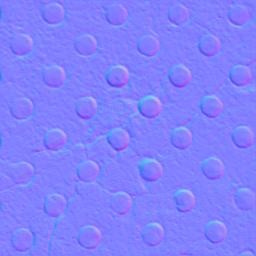} & 
        \includegraphics[width=0.08\linewidth]{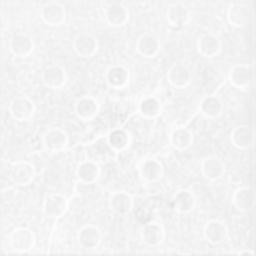} & 
        \includegraphics[width=0.08\linewidth]{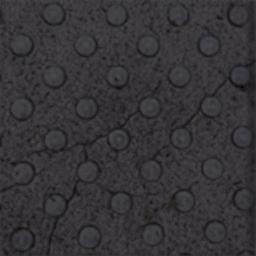} & 
        \includegraphics[width=0.08\linewidth]{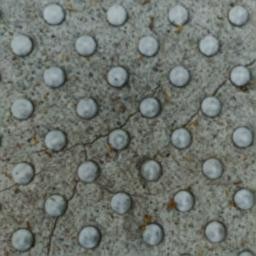} \\

        \vspace{-1mm} \hspace{-1mm} \begin{sideways} \scriptsize{Deschaintre*} \end{sideways}\hspace{0.5mm} & 
        \includegraphics[width=0.08\linewidth]{figures/estim_real/44_real_input.jpg} & 
        \includegraphics[width=0.08\linewidth]{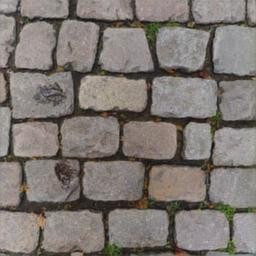} & 
        \includegraphics[width=0.08\linewidth]{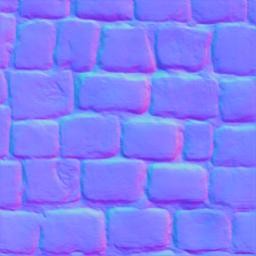} & 
        \includegraphics[width=0.08\linewidth]{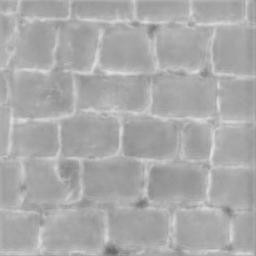} & 
        \includegraphics[width=0.08\linewidth]{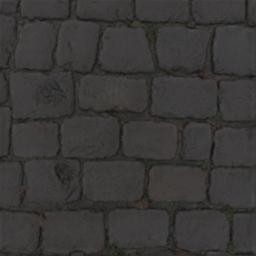} & 
        \includegraphics[width=0.08\linewidth]{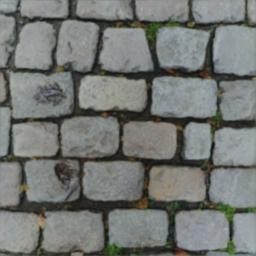} & 
        \hspace{0.25mm} \includegraphics[width=0.08\linewidth]{figures/estim_real/28_real_input.jpg} & 
        \includegraphics[width=0.08\linewidth]{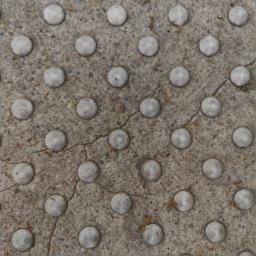} & 
        \includegraphics[width=0.08\linewidth]{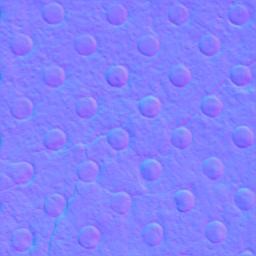} & 
        \includegraphics[width=0.08\linewidth]{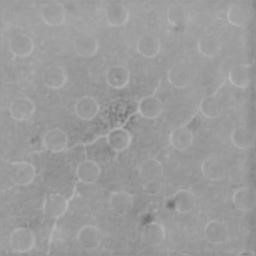} & 
        \includegraphics[width=0.08\linewidth]{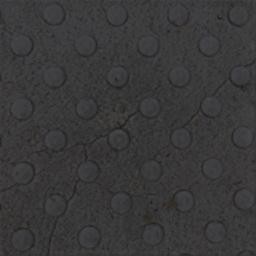} & 
        \includegraphics[width=0.08\linewidth]{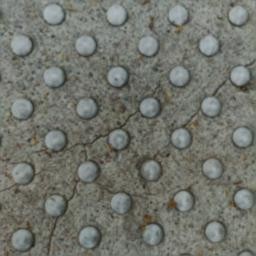} \\

        \vspace{-1mm} \hspace{-1mm} \begin{sideways} \scriptsize{SurfaceNet} \end{sideways}\hspace{0.5mm} & 
        \includegraphics[width=0.08\linewidth]{figures/estim_real/44_real_input.jpg} & 
        \includegraphics[width=0.08\linewidth]{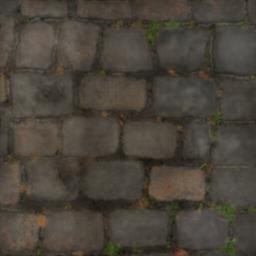} & 
        \includegraphics[width=0.08\linewidth]{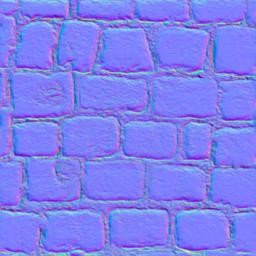} & 
        \includegraphics[width=0.08\linewidth]{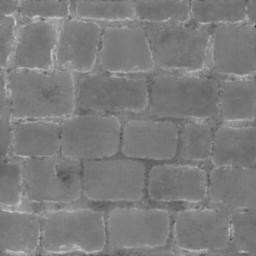} & 
        \includegraphics[width=0.08\linewidth]{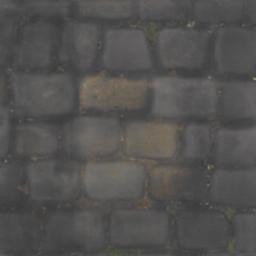} & 
        \includegraphics[width=0.08\linewidth]{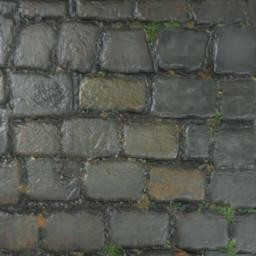} & 
        \hspace{0.25mm} \includegraphics[width=0.08\linewidth]{figures/estim_real/28_real_input.jpg} & 
        \includegraphics[width=0.08\linewidth]{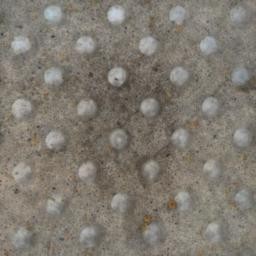} & 
        \includegraphics[width=0.08\linewidth]{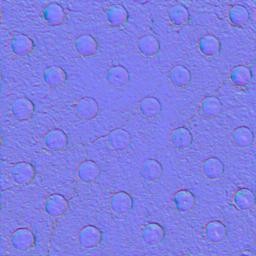} & 
        \includegraphics[width=0.08\linewidth]{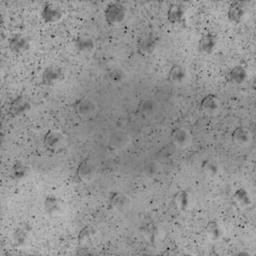} & 
        \includegraphics[width=0.08\linewidth]{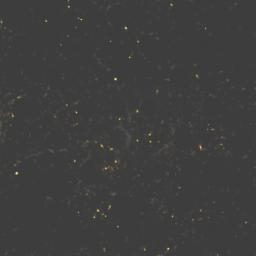} & 
        \includegraphics[width=0.08\linewidth]{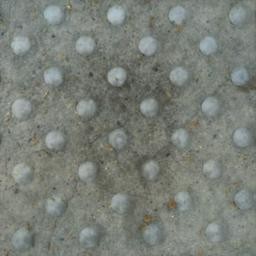} \\

        \vspace{-1mm} \hspace{-1mm} \begin{sideways} \scriptsize{SurfaceNet*} \end{sideways}\hspace{0.5mm} & 
        \includegraphics[width=0.08\linewidth]{figures/estim_real/44_real_input.jpg} & 
        \includegraphics[width=0.08\linewidth]{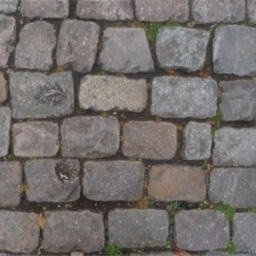} & 
        \includegraphics[width=0.08\linewidth]{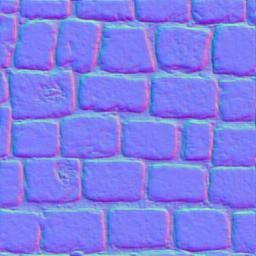} & 
        \includegraphics[width=0.08\linewidth]{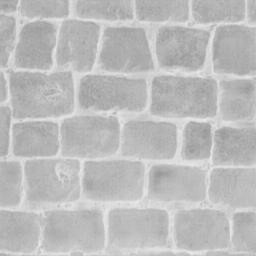} & 
        \includegraphics[width=0.08\linewidth]{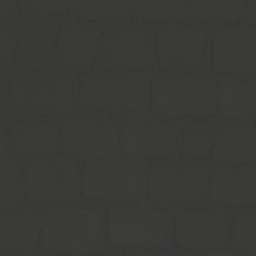} & 
        \includegraphics[width=0.08\linewidth]{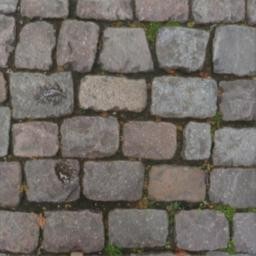} & 
        \hspace{0.25mm} \includegraphics[width=0.08\linewidth]{figures/estim_real/28_real_input.jpg} & 
        \includegraphics[width=0.08\linewidth]{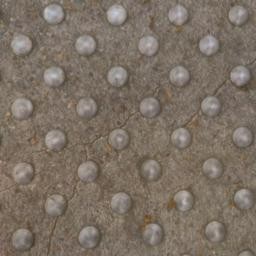} & 
        \includegraphics[width=0.08\linewidth]{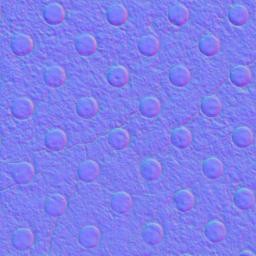} & 
        \includegraphics[width=0.08\linewidth]{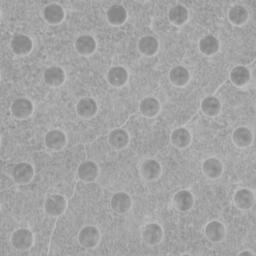} & 
        \includegraphics[width=0.08\linewidth]{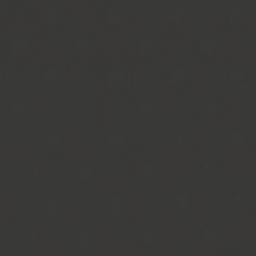} & 
        \includegraphics[width=0.08\linewidth]{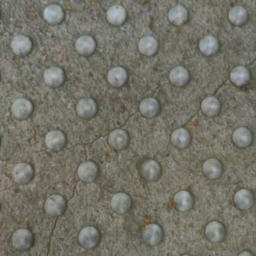} \\
        
    \end{tabular}
    \caption{\textbf{Qualitative material acquisition comparison on real photographs data.} We compare \citet{Deschaintre18} and SurfaceNet~\cite{vecchio2021surfacenet} trained only on \citet{Deschaintre18}'s dataset against the same methods trained on \nameMethod (marked with *). Here the fine-tuned versions better match the input picture, in particular for SurfaceNet.}
    \label{fig:comparison_acquisition_real}
    
\end{figure*}

%% file: figures/generative_results.tex
\begin{figure}
    \centering
    \setlength{\tabcolsep}{.5pt}
    \begin{tabular}{ccccc}
        \multicolumn{5}{c}{MatFuse Retrained} \\
        \vspace{-1mm}\includegraphics[width=0.2\linewidth]{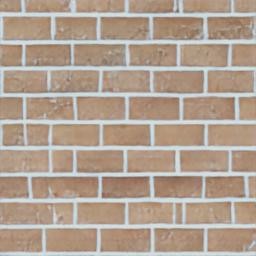} &
        \includegraphics[width=0.2\linewidth]{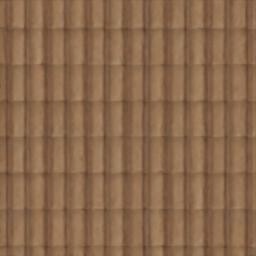} &
        \includegraphics[width=0.2\linewidth]{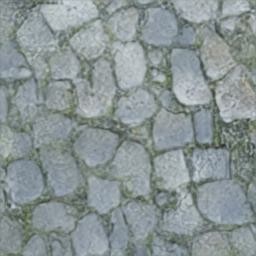} &
        \includegraphics[width=0.2\linewidth]{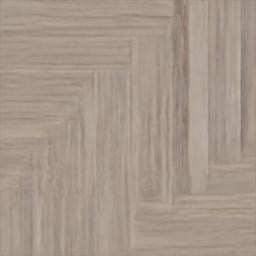} &
        \includegraphics[width=0.2\linewidth]{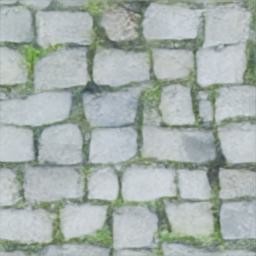} \\
    
        \includegraphics[width=0.2\linewidth]{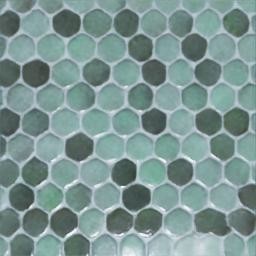} &
        \includegraphics[width=0.2\linewidth]{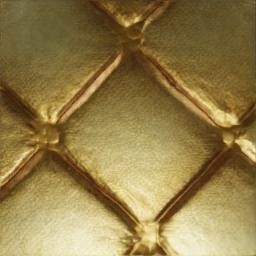} &
        \includegraphics[width=0.2\linewidth]{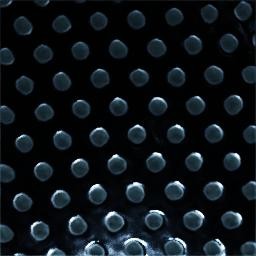} &
        \includegraphics[width=0.2\linewidth]{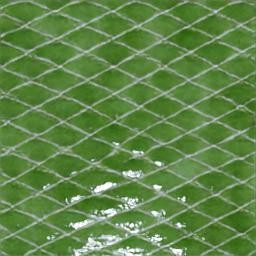} &
        \includegraphics[width=0.2\linewidth]{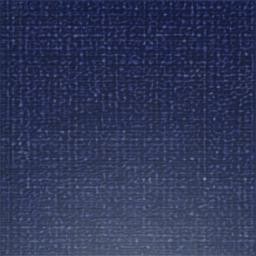} \\
        
        \multicolumn{5}{c}{MatFuse With Fine-tuning} \\
        \vspace{-1mm}\includegraphics[width=0.2\linewidth]{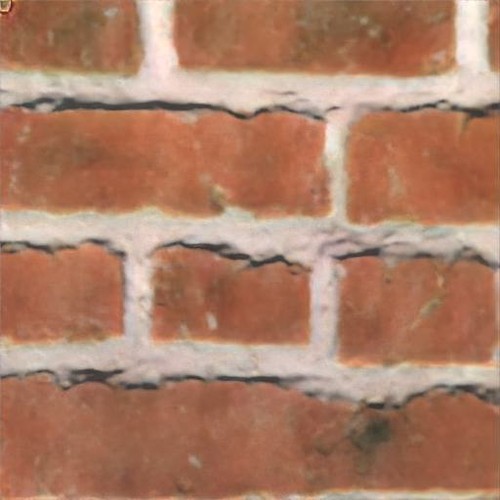} &
        \includegraphics[width=0.2\linewidth]{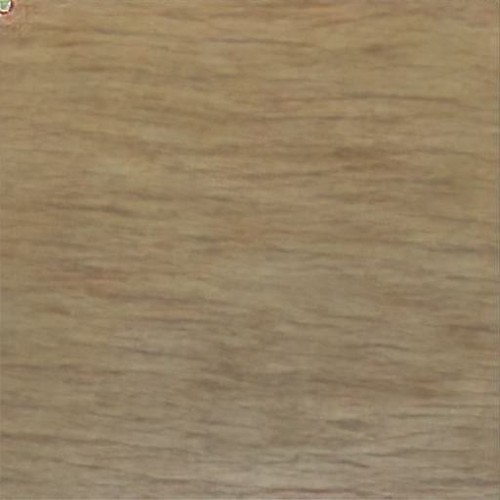} &
        \includegraphics[width=0.2\linewidth]{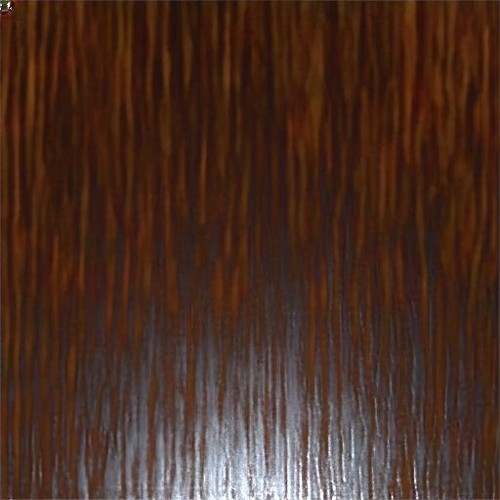} &
        \includegraphics[width=0.2\linewidth]{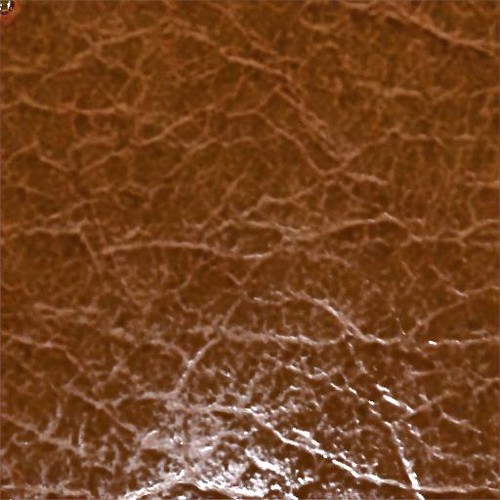} &
        \includegraphics[width=0.2\linewidth]{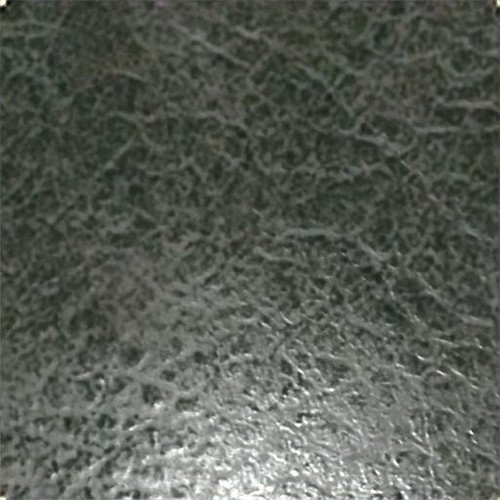} \\
    
        \includegraphics[width=0.2\linewidth]{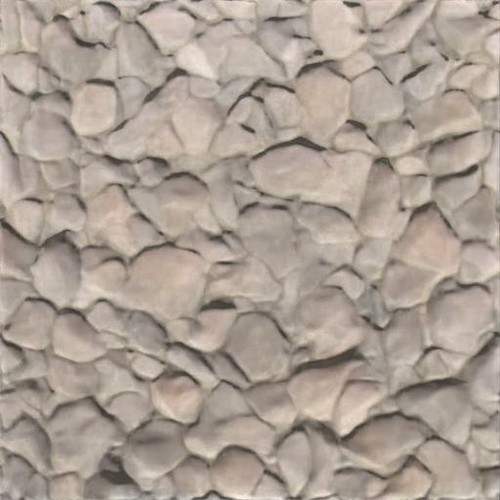} &
        \includegraphics[width=0.2\linewidth]{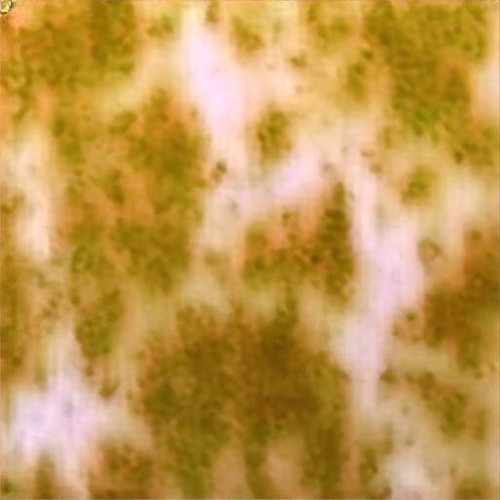} &
        \includegraphics[width=0.2\linewidth]{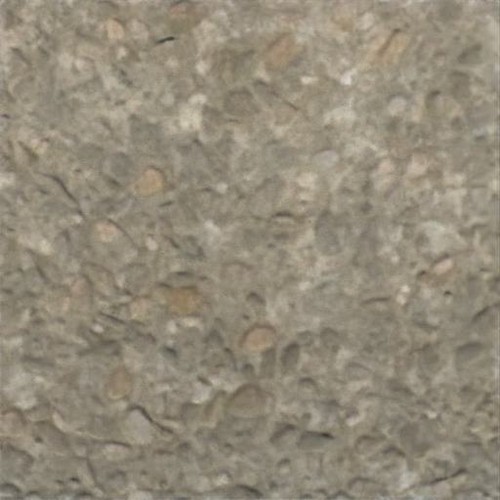} &
        \includegraphics[width=0.2\linewidth]{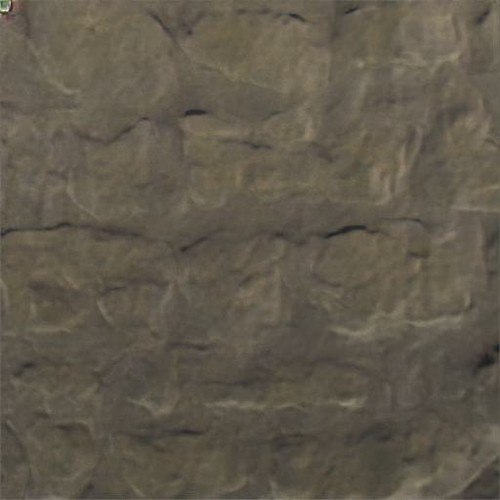} &
        \includegraphics[width=0.2\linewidth]{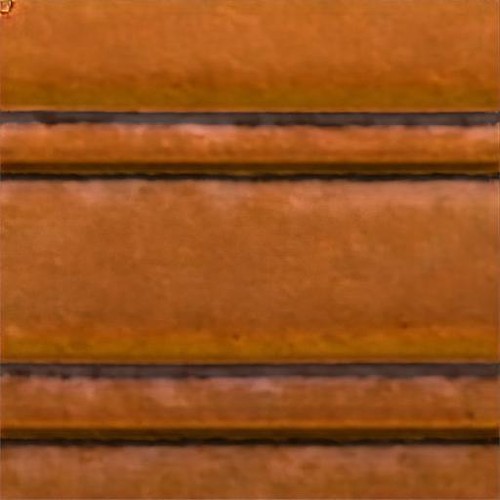} \\

        \multicolumn{5}{c}{MatFuse Without Fine-tuning}\\
        \vspace{-1mm}\includegraphics[width=0.2\linewidth]{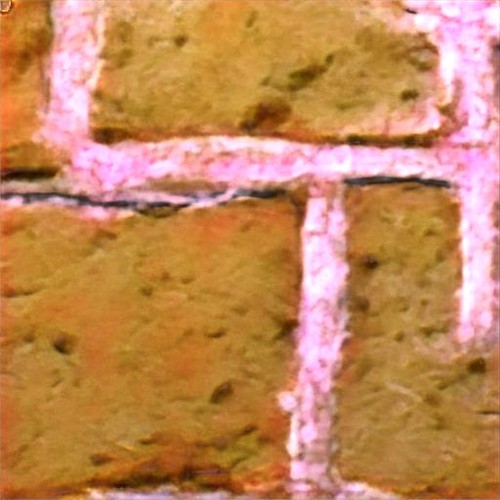} &
        \includegraphics[width=0.2\linewidth]{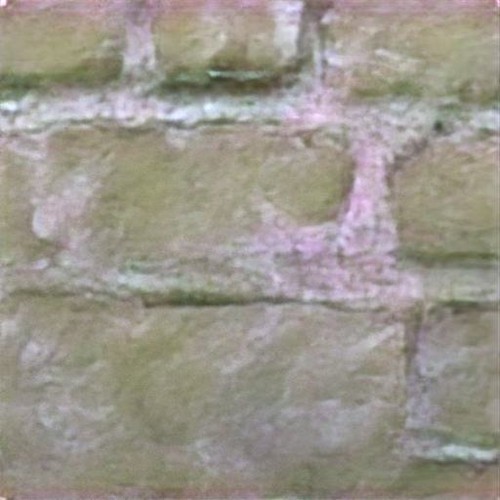} &
        \includegraphics[width=0.2\linewidth]{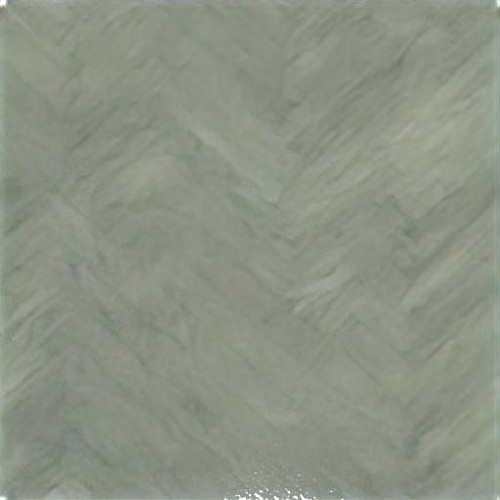} &
        \includegraphics[width=0.2\linewidth]{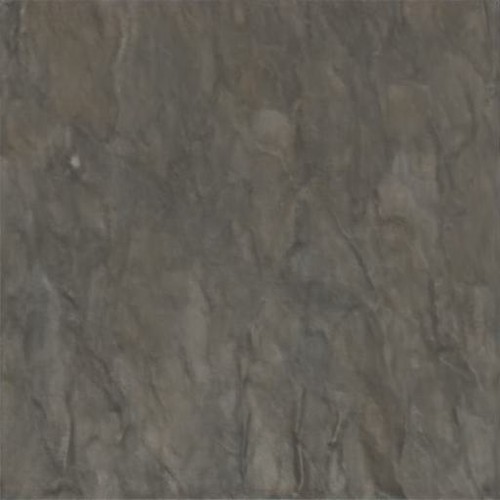} &
        \includegraphics[width=0.2\linewidth]{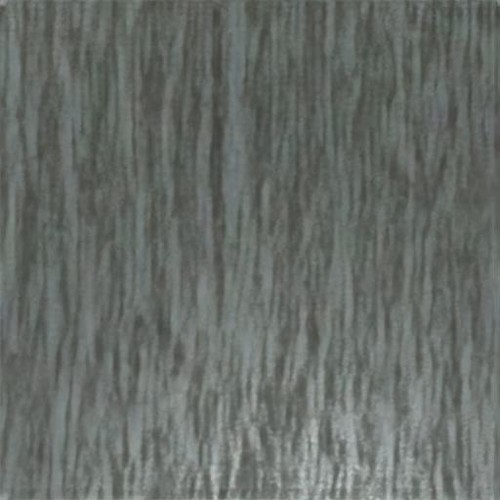} \\
        
        \vspace{-1mm}\includegraphics[width=0.2\linewidth]{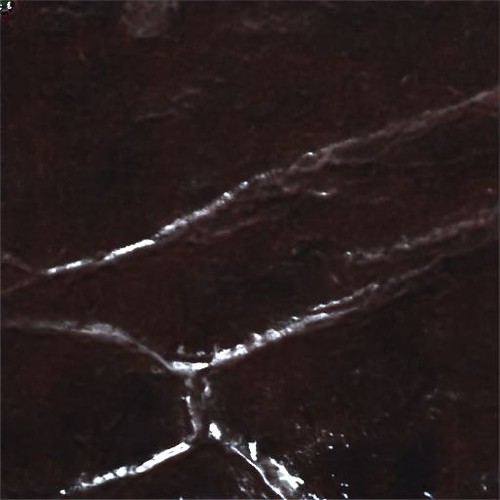} &
        \includegraphics[width=0.2\linewidth]{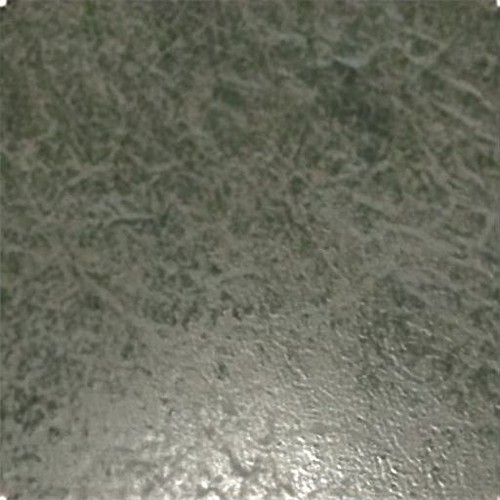} &
        \includegraphics[width=0.2\linewidth]{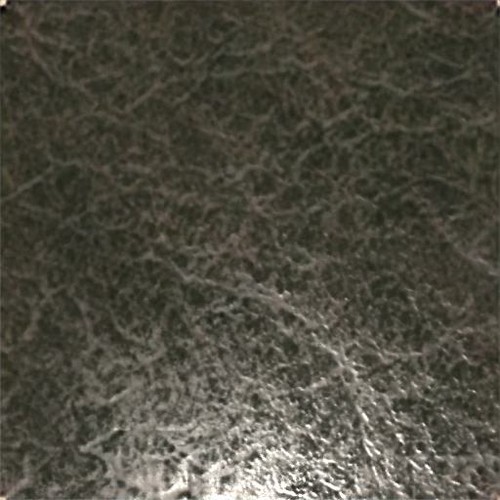} &
        \includegraphics[width=0.2\linewidth]{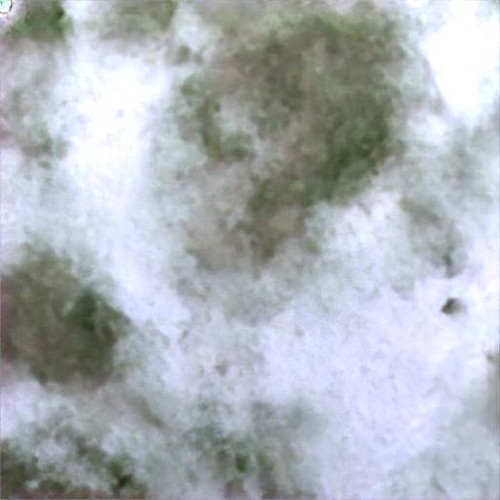} &
        \includegraphics[width=0.2\linewidth]{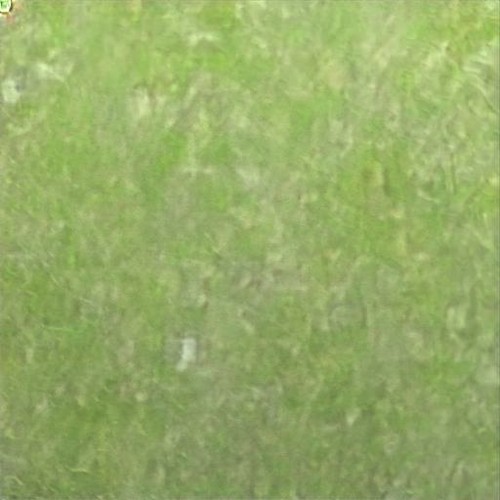} \\

    \end{tabular}
    \vspace{-2mm}
    \caption{\textbf{Comparison on Material Generation.} We show renderings of 10 randomly generated materials, both with and without our dataset. These results are directly sampled from the network and have not been filtered. As expected, additional data help achieve better and more diverse results.}
    \label{fig:generative_comparison}
\end{figure}

%% file: sec/6_discussion.tex
\section{Conclusion}
\label{sec:conclusion}

While there still exists a gap between real materials photographs and rendered synthetic materials, having access to a large, high-quality, database of materials is crucial to facilitate further research around their authoring, understanding, and use in automatic data generation~\cite{vecchio_thesis}. With \nameMethod, we take a significant step to improve material data accessibility, with $4,069$ permissively licensed assets, processed and annotated, alongside millions of augmented renderings. We demonstrate that this additional data immediately improves existing method's acquisition quality and generation diversity. The materials in the dataset are tileable and in 4K resolution, enabling future research on ultra-high resolution for materials. This is of particular interest as most material artists typically work at 4K+ resolutions.

%% file: sec/7_ack.tex
\section{Acknowledgments}
We thank Renato Sortino for providing additional computational power to render the dataset and run trainings.